\documentclass[journal]{IEEEtran}
\usepackage{amsmath,amssymb}
\usepackage{algorithm,algorithmic}
\usepackage{graphicx}
\usepackage{hyperref}
\usepackage{array,booktabs}
\ifCLASSINFOpdf
\else
\fi
 \usepackage[caption=false,font=footnotesize]{subfig}
\begin{document}
%
\title{Semi-supervised Superpixel-based Multi-Feature Graph Learning for Hyperspectral Image Data}
%
%
%

\author{Madeleine~S. Kotzagiannidis, 
        Carola-Bibiane~Sch\"{o}nlieb 
\thanks{M. S. Kotzagiannidis and C.-B. Sch\"{o}nlieb are with the Department
of Applied Mathematics and Theoretical Physics (DAMTP), University of Cambridge, UK. e-mail: mk2060@cam.ac.uk, cbs31@cam.ac.uk. This work has been submitted to the IEEE for possible publication. Copyright may be transferred without notice, after which this version may no longer be accessible.}}
\maketitle

\begin{abstract}
Graphs naturally lend themselves to model the complexities of Hyperspectral Image (HSI) data as well as to serve as semi-supervised classifiers by propagating given labels among nearest neighbours. In this work, we present a novel framework for the classification of HSI data in light of a very limited amount of labelled data, inspired by multi-view graph learning and graph signal processing. 
Given an a priori superpixel-segmented hyperspectral image, we seek a robust and efficient graph construction and label propagation method to conduct semi-supervised learning (SSL). Since the graph is paramount to the success of the subsequent classification task, particularly in light of the intrinsic complexity of HSI data, we consider the problem of finding the optimal graph to model such data. 
Our contribution is two-fold: firstly, we propose a multi-stage edge-efficient semi-supervised graph learning framework for HSI data which exploits given label information through pseudo-label features embedded in the graph construction. 
Secondly, we examine and enhance the contribution of multiple superpixel features embedded in the graph on the basis of pseudo-labels in an extension of the previous framework, which is less reliant on excessive parameter tuning. 
Ultimately, we demonstrate the superiority of our approaches in comparison with state-of-the-art methods through extensive numerical experiments.
\end{abstract}
\IEEEpeerreviewmaketitle
\begin{IEEEkeywords}
Hyperspectral image classification, graph Laplacian learning, semi-supervised learning
\end{IEEEkeywords}

%
\IEEEpeerreviewmaketitle

\section{Introduction}
%
%
%
%
\IEEEPARstart{T}{he} problem of determining the optimal graph representation of a given dataset has been considered for various tasks in different fields, ranging from signal processing to machine learning, yet remains largely unresolved.
Hyperspectral image (HSI) data, with its rich and descriptive spatial and spectral information contained in several hundred bands, encapsulates several layers of dependencies between the high-dimensional pixels and their corresponding labels \cite{guide}, and a plethora of methods have sought to exploit these under different modelling assumptions.\\
In hyperspectral image classification, a classifier is sought to assign a class label to each pixel, in light of arising difficulties including high spectral dimensionality, large spatial variability and limited availability of labels. While the majority of frameworks involve a supervised classifier, due to the time and cost associated with obtaining labelled samples, semi-supervised learning (SSL) has experienced a rapid development by tackling the small sample problem through effective exploitation of both labeled and unlabelled samples \cite{ssl}. \\
Multiple views or features of data can provide rich insights into the underlying data structure, while exploiting diverse information and thus gaining robustness for subsequent tasks. The identification of suitable features, however, represents a problem in itself.  
Inspired by the field of multi-view clustering \cite{Nie1} and in an effort to incorporate more sophisticated model priors, we adopt the perspective of learning a graph from multiple data features (or views) in order to effectively capture the complexity of hyperspectral data, while further integrating the label information into the graph so as to exploit prior knowledge of label dependencies. We further resort to superpixel segmentation of the data in order to effectively reduce the complexity of the classification task, while simultaneously capturing a first level of spectral-spatial dependencies through the clustering into local homogeneous regions. \\
In this paper, we introduce a novel semi-supervised framework for HSI classification which involves the design of a graph classification function which is smooth with respect to both the intrinsic structure of the data, as described via superpixel features, as well as the label space. The proposed graph learning framework is special due to its edge-efficient analytic solution, known to satisfy graph-optimality constraints, and ability to incorporate multiple features, as well as due to its optimization of both the graph and classification function, resulting in pseudo-labels. In addition, in light of the range of parameters required to tune a multi-feature superpixel graph, we propose a variation of the framework which, instead of incorporating pre-set feature weights, learns them by imposing an additional smoothness functional.
We summarize the main contributions as follows:
\begin{itemize}

\item We extend multi-view graph learning to the domain of superpixels and HSI data, with tunable pseudo-label generation incorporated into an updateable graph. In particular, we employ an initial soft graph on which labels are firstly propagated among nearest neighbours to generate pseudo-label features, which are subsequently utilized to inform an improved graph.

\item We propose a pseudo-label-guided framework for HSI feature selection, weighting and subsequent graph construction, enhanced by dynamic pseudo-label-features. To the best of our knowledge, the issue of feature contribution on a multi-feature graph for HSI-data has not yet been tackled, beyond a simple parameter search.

\item We extensively validate our proposed approaches on the basis of three benchmark datasets and demonstrate their superiority with respect to comparable state-of the-art approaches.
\end{itemize}

\noindent The proposed framework facilitates edge-efficient graph and label learning, while flexibly incorporating multiple features which capture spectral-, spatial- and label-dependencies within the given data.
The remainder of this paper is organized as follows: Section II discusses related work, Section III explores the preliminaries, and details of the proposed methods are stated in Section IV. Section V presents the experimental results of our methods in comparison with state-of-the-art approaches, and Section VI contains concluding remarks.

\begin{figure*}[h]
\centering
  \includegraphics[width=5.5in]{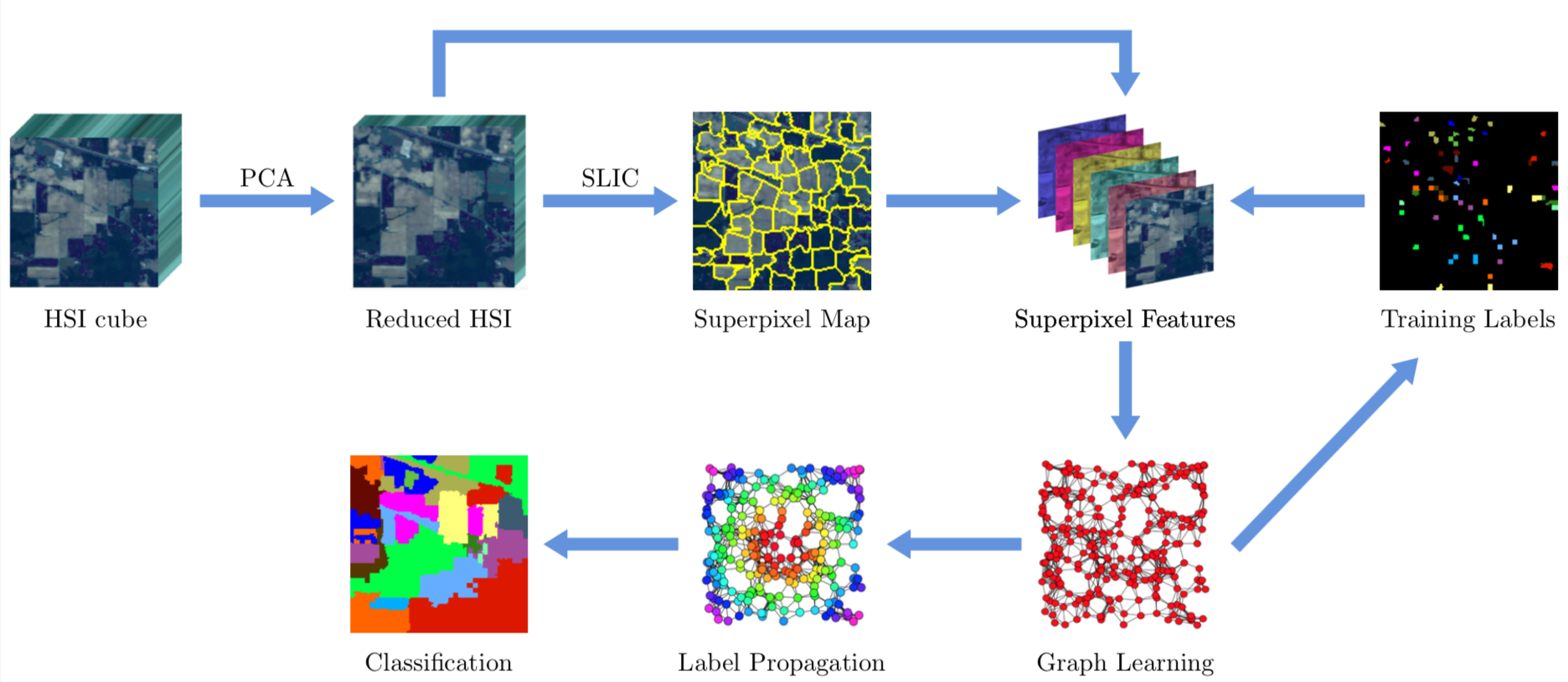}
  \caption{Main Workflow: The proposed MGL and PMGL methods are mainly composed of spectral dimensionality reduction, superpixel segmentation, superpixel feature extraction and superpixel label regularization, followed by a pseudo-label driven graph learning stage, and culminating in the final classification result via label propagation.  }
   \label{fig:w}
\end{figure*}

\section{Related Work}
Works in SSL can be categorized into generative models, which employ probabilistic generative mixture models, co-and self-training methods, low-density separation methods, which seek a decision boundary through low density regions, and graph-based methods \cite{ssl}.
Popular modelling assumptions and concepts have included spectral clustering, data manifolds and local-global consistency \cite{lgc}, according to which nearby points and points residing on the same cluster or manifold are likely to have the same label \cite{ssl}.
In the context of HSI data, earlier approaches comprise kernel methods such as the purely spectral-based SVM \cite{svm} and spectral-spatial multiple kernel learning \cite{sp}. Feature extraction methods, such as \cite{IFRF}, \cite{LBP}, \cite{lcmr}, have sought to characterize a lower dimensional subspace which best captures the spectral-spatial information of the data. Further, classical signal processing concepts such as sparse representation, low-rank, and wavelet analysis have been notably incorporated into SSL methods \cite{lowrank}, \cite{dict}, \cite{guide}.\\
More recent methods have gained in complexity by generating multi-stage workflows with different pre- and post-processing levels as well as by combining the strengths of different classifiers in an effort to
create more sophisticated models which exploit multiple dependencies of HSI data and counteract the small sample problem through a gradual learning process \cite{guide}. \\
Graph-based SSL methods have become increasingly popular due to the superior modelling capabilities of graphs, particularly as a means to counteract the limited amount of labels available, and have included pixel-based \cite{gssl} as well as superpixel-based graph constructions \cite{sp2}, \cite{rlf}, \cite{sperw}, ranging from end-to-end approaches, that utilize graphs for both data modelling and label propagation, to hybrid approaches. We note that many existing methods, such as EPF \cite{epf}, which makes use of the bilateral filter, exhibit underlying graph-like qualities, while not directly or only in part employing graphs, and as a result have exhibited superior performance.
In addition, approaches such as \cite{hong}, have sought to incorporate label information early on in the workflow, known as \textit{pseudo-labelling}, which is progressively refined, in an effort to inform data modelling; nevertheless, the graph construction is not necessarily analytic. While identification of the correct model graph is essential for SSL \cite{graph}, extensive parameter analysis is still inevitable in order to study their influence, as performance is strongly affected by all components of the graph; nevertheless, previous studies have generally not found explorable patterns \cite{graph}, and stability is preferred over excellence in a narrow parameter range.\\
A preceding body of work on multi-view clustering has sought to unify the tasks of data similarity learning (i.e. graph construction) and clustering/label propagation in a joint optimization framework \cite{Nie1}, \cite{multigraph}, \cite{auto}, which, to the best of our knowledge, has not been fully investigated for the joint challenge of HSI data and SSL.
Graph learning and label propagation are separate tasks but recent efforts have opted to combine them, thereby updating and incorporating label information into the graph, under the driving assumption that a single static graph is not sufficient to solve the entire SSL task successfully. Nevertheless, and not least of all for complex data such as HSI, the issue of propagating error noise into each task needs to be addressed. 
There are multiple ways in which the label information can be incorporated into the graph, which include space fusion approaches, i.e. the graph and label space are fused via the addition of a label correlation matrix \cite{DLP} or the removal of differently labelled edges \cite{RMGT}, and implicit approaches, which instead consider the graph as a function of the distances between labels \cite{Nie1}.\\
Deep learning (DL) methods have penetrated graph-based classification in the form of Graph Convolutional Networks (GCNs), which usually require a given pre-constructed graph, such as the spectral-spatial GCN \cite{gcn1}, and have been employed to extract features automatically \cite{zhu}.
Under the assumption that the mapping from the feature to the label space is sufficiently smooth, the theoretical relation between label propagation (LP) and GCNs, as instances which propagate labels and features respectively, has been notably shown to satisfy a smoothness inequality \cite{gcnlpa}. Nevertheless, DL-, and non-DL methods alike, ultimately rely on the optimality of the graph while often performing worse with limited labels. Neural networks have further been employed to learn graphs from scratch (see e.g. \cite{gcn2}), which inevitably comes at a high computational cost, however, for complex and rich datasets such as HSI the formulation of sophisticated (graph) model priors is paramount to the successful performance of any classifier. \\
Graphs ultimately capture model constraints in the form of linear dependencies between data points (as per the graph Laplacian \cite{ansyn}) and, as such, present versatile modelling tools which can help simplify more complex modelling assumptions. Most graph-based methods operate under the assumption that the given graph is optimal and/or that the given data naturally resides on a known graph, so any subsequent tasks and operations are subject to noise pertaining to an imperfect model. Given a data set, approaches have opted to either hand-craft the graph, or learn it automatically through the minimization of a chosen optimization function. In the former case, this has included local and/or adaptive graph weighting and connectivity schemes, which take into account variations in i.a. (spectral) data density, and are increasingly refined but also bear a risk of distorting neighborhood information \cite{spectraldens}. While handcrafted approaches usually lead to higher accuracy, automated ones can be more robust to variable datasets, requiring less parameter tuning while nevertheless being more costly due to lack of an analytic solution. In an effort to combine the advantages of both, we therefore seek an analytic solution to a well-defined optimization problem with the possibility of sophisticated parameter learning (i.e. reduction) through regularization.



\section{Preliminaries and Prior Work: Graph (Laplacian) Learning}
An undirected graph $G=(V,E)$ is characterized by a set of vertices $V$ and a set of edges $E$, and its connectivity is encapsulated in the symmetric adjacency matrix ${\bf W}$, with  $W_{i,j}>0$ if there is an edge between vertices $i$ and $j$, and $W_{i,j}=0$ otherwise. The non-normalized graph Laplacian matrix is defined as ${\bf L}={\bf D}-{\bf W}$, where ${\bf D}=diag({\bf W}{\bf 1})$ is the diagonal degree matrix, with ${\bf 1}$ denoting the vector of 1's. 
The construction of a graph which optimally represents a given dataset ${\mathcal X}=\{{\bf x}_1,...,{\bf x}_N\}$ is a multi-part task, which generally establishes relations of the form $W_{i,j}=A_{i,j} w({\bf x}_i,{\bf x}_j)$ between data points ${\bf x}_i, {\bf x}_j\in \mathbb{R}^d$, where $w:\mathbb{R}^d\times\mathbb{R}^d\rightarrow \mathbb{R}$ is a similarity function and ${\bf A}$ an adjacency matrix, for which $A_{i,j}=1$ iff $j\in\mathcal{N}_i$ for some pre-determined neighborhood $\mathcal{N}_i$ of node $i$, and $A_{i,j}=0$ otherwise. From the smoothness of $w$ to the range of influence $\mathcal{N}$, each component is vital in ensuring effective discriminative data representation.
\\
In graph signal processing, Graph Laplacian Learning (GLL) considers the minimization of the graph Laplacian quadratic form with respect to data (aka graph signal) matrix ${\bf X}$:
\[Tr({\bf X}^T {\bf L} {\bf X})=\frac{1}{2}\sum_{i,j} W_{i,j}||{\bf x}_i-{\bf x}_j||_2^2=\frac{1}{2}||{\bf W}\circ {\bf Z}||_{1,1},\] 
with trace operator $Tr(\cdot)$ and $Z_{i,j}=||{\bf x}_i-{\bf x}_j||_2^2$, and can be alternatively expressed as a weighted sparsity $l_1$-norm of ${\bf W}$, whose minimization enforces connectivity (large $W_{i,j}$) between similar features ${\bf x}_i$ and ${\bf x}_j$. Previous works (see e.g. \cite{kalofolias}, \cite{xdong}), have considered variations of the general framework: 
\begin{equation}\label{eq:opt}
\min_{{\bf W}\in\mathcal{W}}||{\bf W}\circ {\bf Z}||_{1,1}+f({\bf W}),\end{equation}
consisting of the graph Laplacian quadratic form and a, possibly sparsity-promoting, regularization term $f({\bf W})$, subject to graph constraints, e.g. $\mathcal{W}=\{{\bf W}\in\mathbb{R}_{\geq 0}^{N\times N}: {\bf W}={\bf W}^T, diag({\bf W})=0\}$. 
We select $f({\bf W})=\alpha||{\bf W}||^2_F,\alpha\in\mathbb{R}$, with Frobenius norm $||{\bf W}||_F=\sqrt{\sum_{i,j} |W_{i,j}|^2}$, which controls sparsity by preventing the occurrence of strong edges, and due to the edge-locality of the functional, facilitates the decomposition of the problem into a sum over graph edges, and hence an analytic solution.\\
Specifically, the optimization problem of Eq. (\ref{eq:opt}) becomes separable for $f({\bf W})=\alpha||{\bf W}||^2_F$, and can be rewritten in row-wise form \cite{Nie1}, for row ${\bf W}_i$ of ${\bf W}$, as
\begin{equation}
\label{eq:pro}
\min_{{\bf W}_i{\bf 1}=1, {\bf W}_i\geq {\bf 0}, W_{i,i}=0} \sum_{j=1}^N Z_{i,j} {W}_{i,j} +\alpha\sum_{j=1}^N W_{i,j}^2
\end{equation}
yielding the closed-form solution
\[{W}_{i,j}=\left(\eta_i-\frac{Z_{i,j}}{2\alpha}\right)_{+},\]
where $(x)_{+}=\max (0,x)$ and scalar $\eta_i$. Since the second term creates a dense edge pattern, one can further enforce kNN connectivity by determining the maximal $\alpha_i$ per row s.t. the optimal ${\bf W}_i$ has exactly $k$ non-zeros (see \cite{Nie1} for details). This leads to
 \begin{equation}
 \label{eq:gg}
 W_{i,j}=
\begin{cases}
\frac{Z_{i,k+1}-Z_{i,j}}{k Z_{i,k+1}-\sum_{h=1}^k Z_{i,h}}, & j\leq k\\
0, & j> k.
\end{cases}
\end{equation}
where entries $\{Z_{i,1},...,Z_{i,N}\}$ are assumed to be ordered from small to large wlog, and we have $\eta_i=\frac{1}{k}+\frac{\sum_{h=1}^k Z_{i,h}}{2k\alpha_i}$ and $\alpha_i=(k/2) Z_{i,k+1}-(1/2)\sum_{h=1}^k Z_{i,h}$. Here, $W_{i,i}=0$ is enforced and $Z_{i,i}=0$ appended at the end. Symmetrization is achieved through ${\bf L}={\bf D}-\frac{{\bf W}+{\bf W}^T}{2}$. \\
It has been noted that this approach is computationally efficient due to its analytic solution and in-built sparsity which does not require oblique tuning of $\alpha$ (instead only requiring the straight-forward number of edges $k$) and is further scale-invariant w.r.t. feature vectors ${\bf x}_i$.\\
\\
For different features (views) of type $v$, we consider $Z_{i,j}=\sum_v c_v Z_{i,j}^v$ with feature coefficients $c_v\in\mathbb{R}$. In order to reduce parameters, it has been proposed to constrain the  coefficients to be proportional to pre-set feature-dependent functionals and subject to regularization \cite{multigraph}, \cite{auto}.

\section{Proposed Method: Superpixel-based Multi-feature Graph Learning}

In the proposed method, several instances of data dependencies are exploited in a multi-level workflow: after conducting an initial spectral dimensionality reduction using PCA \cite{pca}, we consider a priori the segmentation of the hyperspectral image into superpixels to define local regions of homogeneous spectral content and to reduce spatial dimensionality. Subsequently, we compute analytic superpixel features, which capture different image properties, and extrapolate given pixel labels to superpixel labels via a simple averaging filter. We then compute an initial graph $G_0$ and form pseudo-label features through a soft label propagation to nearest neighbors on $G_0$. This is subsequently updated and refined, before a final graph classifier is applied (see Fig. \ref{fig:w}). 
\subsection{Superpixel-based Feature Extraction}
\label{sp1}
Given the raw HSI data cube ${\bf I}\in\mathbb{R}^{X\times Y\times B}$, we apply dimensionality reduction in the first instance in the spectral domain using PCA to obtain the reduced $\tilde{{\bf I}}\in\mathbb{R}^{X\times Y\times b},\ b<<B$. \\
Subsequently, the first PC component is used to conduct superpixel segmentation via SLIC \cite{slic}, resulting in the 2D superpixel labelling map $\tilde{{\bf S}}\in\mathbb{R}^{X\times Y}$ with $N$ superpixels:
\[{\bf S}_k \ \text{s.t.}\ {\bf S}_k=\{{\bf S}_{(i,j)} | {\bf S}_{(i,j)}=k\}, \ \tilde{{\bf S}}=\cup_{k=1}^N{\bf S}_k.\]
While it has been established that the goodness and scale of the superpixel segmentation is foundational for the success of subsequent data modelling and classification tasks, it is not the objective of this work to optimize this particular instance of the workflow; as such, we select SLIC \cite{slic} as the base superpixel segmentation algorithm and determine the number of superpixels approximately according to \cite{colsup}, which takes into account both the size and resolution of the HSI image (albeit not the scene complexity).\\
Let ${\bf Y}\in\mathbb{R}^{XY\times c}$ denote the initial class indicator matrix with $Y_{i,j}=1$ if pixel $i$ belongs to class $j$ for $c$ classes. Following superpixel segmentation, we regularize this to ${\bf Y}^S\in\mathbb{R}^{N\times c}$ by averaging over the existing pixel labels per superpixel. Specifically, $Y^S_{i,j}$ records the number of pixels per superpixel ${\bf S}_i$ which belong to class $j$ divided by the total number of pixels in ${\bf S}_i$.
\\
As statistical descriptors for the superpixels, we consider the features as proposed in \cite{sp}, comprising the mean feature vector ${\bf s}_k^M$ 
\[{\bf s}_k^M=\frac{\sum_{i,j}^{N_{k}} \tilde{{\bf I}}_{(i,j)}}{N_k}, \ {\bf S}_{(i,j)}=k,\ k=1,...,N\]
which takes a simple average of the $N_k$ pixels per superpixel $k$, and the spatial-mean feature vector ${\bf s}_k^S$
\[{\bf s}_k^S=\sum_{i=1}^J w_{k,a_i} {\bf s}_{a_i}^M, \ w_{k,a_i}=\frac{\exp{(-||{\bf s}^M_{a_i}-{\bf s}^M_k ||_2^2/h)}}{\sum_{i=1}^J \exp{(-||{\bf s}^M_{a_i}-{\bf s}^M_k ||_2^2/h)}}\]
which constitutes a weighted sum of the mean feature vectors of adjacent superpixels, given by index set $\mathcal{A}_k=\{a_1, ...,a_J\}$ for the $k$-th superpixel and with pre-set scalar $h\in\mathbb{R}$.
Further, we consider the centroidal location of each superpixel as:
\[{\bf s}_k^C=\frac{\sum_{j=1}^{N_k} l_{k,j}}{N_k}\]
where $l_{k,j}$ denotes the 2D image coordinate of the $k$-th superpixel. 
We note that an optimized graph can only be as good as the extracted features it is built upon, however, the task of feature optimization, in line with superpixel segmentation, represents a problem in itself and is not the focus of this work.
\subsection{Dynamic Graph Learning and Label Propagation}
In the following, we wish to learn a superpixel HSI graph and conduct classification through label propagation. \\Consider the joint optimization problem over the graph ${\bf W}$ and the labelling function ${\bf F}$ 
\begin{equation}
\label{eq:op}
\min_{{\bf W}, {\bf F}}\sum_{i,j} (Z_{i,j}+\gamma Z_{i,j}^F)W_{i,j}+\alpha W_{i,j}^2,\end{equation}\[ \ \text{s.t. }\ \sum_j W_{i,j}=1,\ W_{i,j}\geq 0, \ {\bf F}_l={\bf Y}^S_l.\]
with $Z_{i,j}^F=||{\bf f}_i-{\bf f}_j||_2^2$, where ${\bf f}_i$ denotes the $i$-th row of ${\bf F}$, and ${\bf Y}^S_l\in\mathbb{R}^{l\times c}$ is the labelled submatrix of ${\bf Y}^S$.
Via alternating optimization, the optimal graph ${\bf W}$ can be computed according to Eq. (\ref{eq:gg}) with $Z_{i,j}\rightarrow Z_{i,j}+\gamma Z_{i,j}^F$ and by absorbing $\alpha$ into $k$, the number of edges per row, while for ${\bf F}$, this yields the solution 
\begin{equation}
\label{eq:sol}
{\bf F}_u=-{\bf L}_{u,u}^{-1} {\bf L}_{u,l}{\bf Y}^S_l
\end{equation}
for the unlabelled superpixel nodes, which is also known as harmonic label propagation \cite{auto}, \cite{harmonic}. Here, ${\bf L}_{u,u}$ denotes the graph Laplacian submatrix with rows and columns corresponding to unlabelled nodes.
The final superpixel labels are assigned via the decision function \begin{equation}\label{eq:final}
y_i=\arg \max_j F_{i,j},\ \ \forall j=1,..., c.
\end{equation}
Accordingly, we require the optimal graph to exhibit smoothness with respect to both the pre-designed superpixel features, as extracted from the HSI data, as well as the labelling function ${\bf F}$. The functional $Tr({\bf F}^T{\bf L}{\bf F})=\frac{1}{2}\sum_{i,j}Z_{i,j}^F W_{i,j}$, given ${\bf L}$, has been employed as a standalone graph classifier (e.g. \cite{lgc}, \cite{harmonic}), while in the present framework, it is further utilized to enrich the graph construction process, rendering it dynamic.\\
\\
Let ${\bf Z}^v$ with $Z^v_{i,j}=||{\bf s}^v_i-{\bf s}_j^v||^2_2$ denote the Euclidean distance matrix between superpixel feature vectors of type (or view) $v\in\{M,S,C\}$, as previously defined in Sect.\ \ref{sp1}, and $c_v$ the feature weight. 
In a variation of the above, we propose to employ the superpixel multi-feature dynamic graph $G$ of the basic form as in Eq. (\ref{eq:gg}) with weighted pairwise distances $Z_{i,j}=\sum_v c_v Z^v_{i,j}$ and $Z_{i,j}^{\tilde{F}}=||\tilde{{\bf f}}_i-\tilde{{\bf f}}_j||_2^2$, the latter of which we define as \textit{pseudo-label features} 
\[\tilde{{\bf f}}_i[j]=\sum_k{ P}_{i,k}^0{Y}_{k,j}^S,\ j=1,...,c \] 
where $\tilde{{\bf f}}_i$ denotes the $i$-th row of $\tilde{{\bf F}}\in\mathbb{R}^{N\times c}$, ${\bf W}_0$ the initial graph based on $Z_{i,j}$ and ${\bf P}^0={\bf D}^{-1}_0{\bf W}_0$ its random-walk normalized version. In particular, $\tilde{{\bf F}}$ constitutes one instance of a random walk, as opposed to the fully converged solution in Eq. (\ref{eq:sol}). The motivation behind this construction is to provide a soft pre-labelling approach by propagating given labels among their nearest neighbors, as determined through an initial superpixel graph ${\bf W}_0$ based solely on HSI superpixel features, thereby merging information from the label and superpixel dependencies. The resulting distance matrix ${\bf Z}^{\tilde{F}}$ is then utilized as an additional component to rebuild the graph, with large values penalizing nodes not in the same class, further rendering the graph construction dynamic and implicit. Final label propagation on this graph is conducted via the converged harmonic solution in Eq. (\ref{eq:sol}) and class assignment via Eq. (\ref{eq:final}). We summarize the approach in Algorithm 1.
\begin{algorithm}
\caption{Multi-Feature Graph SSL for HSI Data}
\begin{algorithmic}[1]
  \STATE {\bf INPUT:} raw HSI cube ${\bf I}$, label matrix ${\bf Y}$, parameters: k, $\{c_v\}_v$, $\gamma$.
  \STATE {\bf OUTPUT:} Classification map ${\bf F}$.
  \STATE Apply PCA on ${\bf I}$ to obtain $\tilde{{\bf I}}$.
  \STATE Conduct superpixel segmentation to obtain $\tilde{{\bf S}}$ and label regularization to obtain ${\bf Y}^S$.
  \STATE Extract superpixel features $\{{\bf s}^M,{\bf s}^S,{\bf s}^C\}$.
  \STATE Compute initial superpixel-feature graph ${\bf W}_0$ with $Z_{i,j}=\sum_v c_v Z_{i,j}^v$, pre-set $c_v$, $v\in\{S,M,C\}$, via Eq. (\ref{eq:gg}) $\&$ symmetrize. 
  \STATE Compute pseudo-label features $\tilde{{\bf F}}={\bf P}^0{\bf Y}^S$ and ${\bf Z}^{\tilde{F}}$.
  \STATE Update graph with $\tilde{Z}_{i,j}=Z_{i,j}+\gamma Z_{i,j}^{\tilde{F}}$ and symmetrize.
  \STATE Compute unknown labels with graph classifier ${\bf F}_u=-{\bf L}_{u,u}^{-1} {\bf L}_{u,l}{\bf Y}^S_l$. 
  \STATE Assign final classes via Eq. (\ref{eq:final}).
\end{algorithmic}
\end{algorithm}
\\
\textit{Remark:} The RBF kernel with $W_{i,j}=\exp\left(-\frac{||{\bf x}_i-{\bf x}_j||_2^2}{\sigma^2}\right)$ constitutes a prominent approach to model graph weights \cite{graph} and is incidentally the result of the optimization problem in Eq. (\ref{eq:opt}) with $f({\bf W})=\sigma^2\sum_{i,j}W_{i,j}\log W_{i,j}$ in normalized form. Nevertheless, performance is strongly affected by the tuning of $\sigma$ and graph connectivity, the latter of which is not embedded into the graph solution and both of which are non-trivial. \\
While the proposed approach simplifies the issue of tuning edge connectivity and neighbourhood range, thus increasing robustness, it still comprises a range of parameters, which are categorized into: $k$ (the number of nearest neighbour edges), $\{c_v\}_v$ (superpixel feature weights), and $\gamma$ (pseudo-label feature weight). 
As the goodness of the graph is dependent on the goodness of its features, we proceed to simultaneously learn feature weights and update the graph with the goal to help guide as well as minimize uninformed parameter-tuning.

\subsection{Parameter-optimal Multi-feature Graph Learning}

While a reduction of parameters generally occurs at the sacrifice of performance and cannot replace a thorough parameter search, we investigate the possibility of a pseudo-label guided parameter reduction and further propose a variation of the preceding framework, inspired by \cite{multigraph}, in an effort to facilitate training and generalizability to diverse datasets.\\
Consider individual superpixel feature graphs, denoted with ${\bf A}^v$, whose entries are computed from $Z_{i,j}^v$ using Eq. (\ref{eq:gg}) with $k$ edges per row, and initialize the global graph ${\bf W}=\sum_{v=1}^V c_v {\bf A}^v$ with $c_v=\frac{1}{V}$. We assume that the deviation $r^v=||{\bf W}-{\bf A}^v||_F^2$ from ${\bf W}$ is inversely related to the feature importance $c_v$.  After an initial pseudo-label computation $\tilde{{\bf F}}$ using Eq. (\ref{eq:sol}), we update the global graph ${\bf W}$ by solving
\begin{equation}
\label{eq:gl}
\min_{{\bf W}_i\geq 0, {\bf W}_i {\bf 1}=1}\left|\left|{\bf W}_i+\frac{(\gamma_1/2) {\bf Z}_{i}^{WF}-\sum_v c_v {\bf A}_i^v}{\sum_v c_v}\right|\right|_2^2\end{equation}
In contrast to the previous method, we determine the pseudo-labels $\tilde{{\bf F}}$ here via the converged harmonic solution in Eq. (\ref{eq:sol}) and apply a binary mask ${W}$ which holds the nonzero locations of the current global graph estimate ${\bf W}$, giving the Hadamard product ${\bf Z}^{WF}={W}\circ {\bf Z}^{\tilde{F}}$. As such, when solving Eq. (\ref{eq:gl}) the positions of the non-zero graph weights remain fixed, while their values are perturbed, i.e. re-weighted or eliminated according to pseudo-label information. This prevents the formation of noisy edges when ${\bf Z}^{\tilde{F}}$ is dense, and constitutes an alternative to soft pseudo-labelling (for which we previously employed one random walk). Subsequently, we regularize the weights $c_v$ via an $l_2$-norm term:
\begin{equation}
\label{eq:coef}
\min_{{\bf c}} \sum_v c_v ||{\bf W}-{\bf A}^v||_F^2+\gamma_2 ||{\bf c}||_2^2,\ \text{s.t.}\ c_v\geq 0, \ {\bf c}^T{\bf 1}=1\end{equation}
which can be simplified to 
\begin{equation}
\label{eq:coef2}\min_{c_v \geq 0, {\bf c}^T{\bf 1}=1}\left|\left|{\bf c}+\frac{{\bf r}}{2\gamma_2}\right|\right|_2^2.\end{equation}
Notably, Eq. (\ref{eq:pro}) can be written in the same form as Eqs. (\ref{eq:gl}) and (\ref{eq:coef2}), which constitute the Euclidean projection onto the probabilistic simplex \cite{simplex}; however, as we cannot apply the same kNN simplification, we solve the latter two iteratively, whereby we employ Newton's method to enforce the unity sum constraint. As such, both the graph edge learning and feature weight learning stage are essentially the same.
\\
By recomputing ${\bf Z}^{WF}$ and then ${\bf W}$ with corresponding parameter $\gamma_3$, we obtain a pseudo-label enhanced graph which is used by the graph classifier of Eq. (\ref{eq:sol}) to obtain the final solution.\\
Overall, $\gamma_2$ controls the disparity between feature weights, while replacing $V$ parameters with one, while $\gamma_1$ and $\gamma_3$ regulate the pseudo-label contribution at different stages. Here we employ pseudo-labels in a two-fold way to inform feature contribution as well as to form a separate feature embedded in the graph. While this can be tuned with a single parameter $\gamma_1$, in practice, we observe that performance benefits from weighting the steps separately by introducing $\gamma_3$, as we will demonstrate in Sect. V. 
We summarize the graph learning stage of the approach in Algorithm 2; here, each computed graph is a posteriori symmetrized via $\frac{{\bf W}+{\bf W}^T}{2}$. The approach is similar to solving the joint optimization problem of Eqs. (\ref{eq:op}) (with respect to ${\bf F}$) and (\ref{eq:coef}) alternately, as each step constitutes an optimal closed-form solution; however, we refrain from further iterations between steps ${\bf 4}$ and ${\bf 6}$ to limit possible noise resulting from pseudo-labelling.\\
We note that large deviations in (feature) scales, and thus in ${\bf r}$, result in binary weights (i.e. single-feature selection), which can be remedied, in part, by tuning $\gamma_2$, as well as by refining the feature selection. For this approach, we introduce two composite superpixel feature measures, which merge the centroidal feature, which is less informative for a standalone feature graph, with either of the spectral-content features. Specifically, we consider the multiplicative $Z^v_{i,j}\circ Z_{i,j}^C$ and additive $Z_{i,j}^v+\lambda Z_{i,j}^C, v\in\{M, S\}$, with $v$ chosen as per dataset and $\lambda \sim \sigma_v/\sigma_C$, where $\sigma_v=\sum_{i,j} Z_{i,j}^v/N^2$ denotes the scale per feature. 
\begin{algorithm}
\caption{Parameter-optimal Multi-Feature Graph SSL}
\begin{algorithmic}[1]
  \STATE Initialize superpixel-feature graphs ${\bf A}^v$ with ${\bf Z}^v$ via Eq. (\ref{eq:gg}) and ${\bf W}^0=\sum_v c_v {\bf A}^v$ with $c_v=\frac{1}{V}$, then symmetrize.
  \STATE Compute pseudo-label features $\tilde{{\bf F}}$ via Eq. (\ref{eq:sol}) and ${\bf Z}^{WF}=\textit{W}\circ{\bf Z}^{\tilde{F}}$.
  \STATE Update graph with $\tilde{Z}_{i,j}=(\frac{\gamma_1}{2} Z_{i,j}^{WF}-\sum_v c_v A^v_{i,j})/\sum_v c_v$ with pre-set $\gamma_1$ by solving Eq. (\ref{eq:gl}), symmetrize.
   \STATE Compute feature weights $c_v$ with pre-set $\gamma_2$ via Eq. (\ref{eq:coef2}). 
   \STATE Update pseudo-labels $\tilde{{\bf F}}$ via Eq. (\ref{eq:sol}). 
   \STATE Update graph with pre-set pseudo-label weight $\gamma_3$ in Eq. (\ref{eq:gl}), then symmetrize.
  \STATE Compute final labels via graph classifier ${\bf F}_u=-{\bf L}_{u,u}^{-1} {\bf L}_{u,l}{\bf Y}^S_l$ and Eq. (\ref{eq:final}). 
\end{algorithmic}
\end{algorithm}\\
\noindent \textit{Remark}: One could further consider the constraint $r^v=\sum_{i,j} Z^v_{i,j} W_{i,j}$ in Eq. (\ref{eq:coef}) as a non-separable way to estimate the feature weights, however, we observe that discrepancies in scaling render the parameter $\gamma_2$ more difficult to tune. Instead, we opt to separate the graph into the sum of individual feature graphs. While this bears the bias of reduced global interaction between the features (i.e. in Eq. (\ref{eq:gg}) the sum $Z_{i,j}=\sum_{v} c_vZ^v_{i,j}$ drives the assignment of the nearest $k$ edges), we remedy this by incorporating pseudo-labels into the framework as a means to perturb the solution and reinforce inter-and intra-class relations as well as by introducing composite features.

\section{Experimental Results}
\subsection{Dataset Description}
We validate our approach on three benchmark HSI datasets:\\
{\bf Indian Pines:} This data set was gathered by an Airborne Visible/Infrared Imaging Spectrometer (AVIRIS) over an agricultural site in Indiana and consists of $145\times 145$ pixels with a spatial resolution of 20 m per pixel. The AVIRIS sensor has a wavelength range from $0.4$ to $2.5$ $\mu m$, which is divided into 224 bands, of which 200 are retained for experiments. There are 16 classes, the distribution of which is imbalanced, with Alfalfa, Oats and Grass/Pasture-mowed containing relatively few labeled samples.\\
{\bf Salinas:} The second data set was similarly acquired by the AVIRIS sensor over Salinas Valley, California, comprising $512\times 217$ pixels with a notably higher spatial resolution of $3.7$ m per pixel. Further, 204 bands are retained. The scene contains 16 classes, covering i.a. soils and fields.\\
{\bf University of Pavia:} The third data set was collected by the Reflective Optics System Imaging Spectrometer (ROSIS), containing $610\times 340$ pixels with a spatial resolution of $1.3$ m per pixel, the highest of the three datasets. The spectral range from $0.43$ to $0.86$ $\mu m$ is divided into 115 spectral bands, of which 103 are retained. The urban site contains 9 classes and covers the Engineering School at the University of Pavia.
\subsection{Experimental Design}
In the following, we conduct the experimental evaluation of our method on the benchmarking datasets and demonstrate its superiority compared to state-of-the-art algorithms. For evaluation, experiments are repeated 10 times and performance is assessed on the basis of the average and standard deviation of three quality indeces: the overall accuracy (OA), as the percentage of correctly classified pixels, the average accuracy (AA), as the mean of the percentage of correctly classified pixels per class, and the Kappa Coefficient ($\kappa$), as the percentage of correctly classified pixels corrected by the number of agreements expected purely by chance. 
Further, we compare performance with the Local Covariance Matrix Representation (LCMR) \cite{lcmr}, the Edge-Preserving Filtering (EPF) \cite{epf}, the Image Fusion and Recursive Filtering (IFRF) \cite{IFRF}, and the SVM \cite{svm} methods, which, as established state-of-the-art methods, were specifically chosen as comparisons due to their inherent spectral-spatial modelling techniques (with exception of the purely spectral SVM). The SVM algorithm is implemented in the LIBSVM library \cite{libsvm}, adopting the Gaussian kernel with fivefold cross validation for the classifier. We further adopt model-parameters as specified in these works. 
The proposed methods are abbreviated as Multi-Feature Graph Learning (MGL) and Parameter-optimal Multi-Feature Graph Learning (PMGL) respectively.

\subsection{Parameter Specification}
For both proposed methods, we conduct PCA on the standardized data to explain $99.8\%$ of the data's variance and employ SLIC \cite{slic} for superpixel segmentation with a compactness of $10$, where we fix the number $K$ of superpixels per dataset to $K=1287$ (Indian Pines), $K=2237$ (Salinas) and $K=3080$ (University of Pavia). For all graph constructions, we set $k=10$ as the number of edges per node, and for the spatial mean feature construction, we select $h=15$.\\
For the first method (MGL), we employ pseudo-label feature weight $\gamma=10$ and superpixel feature weights $c_S=1$, $c_M=0.5$, $c_C=10^{-2}$ (Indian Pines), $c_S=0.5$, $c_M=5$, $c_C=10^{-5}$ (University of Pavia), and $c_S=1$, $c_M=0.1$, $c_C=10^{-4}$ (Salinas).\\
For the second method (PMGL), we employ $\gamma_1=0$, $\gamma_2=30$, $\gamma_3=1$ (Indian Pines), $\gamma_1=20$, $\gamma_2=40$, $\gamma_3=0$ (University of Pavia) and $\gamma_1=1$, $\gamma_2=30$, $\gamma_3=1$ (Salinas).
Further, for the latter method we construct three individual feature graphs respectively based on the following feature distances: $\{{\bf Z}^M, {\bf Z}^S, {\bf Z}^C\circ {\bf Z}^S\}$ (Indian Pines), $\{{\bf Z}^M, {\bf Z}^S, {\bf Z}^M+\lambda {\bf Z}^C\}$ (University of Pavia), $\{{\bf Z}^S, {\bf Z}^M+\lambda {\bf Z}^C, {\bf Z}^S+\lambda {\bf Z}^C\}$ (Salinas), which were deemed to summarize most effectively the main properties of the different HSI scenes.

\subsection{Experimental Results \& Discussion}
\begin{figure*}[!t]
\subfloat[Indian Pines]{  \includegraphics[width=2.2in]{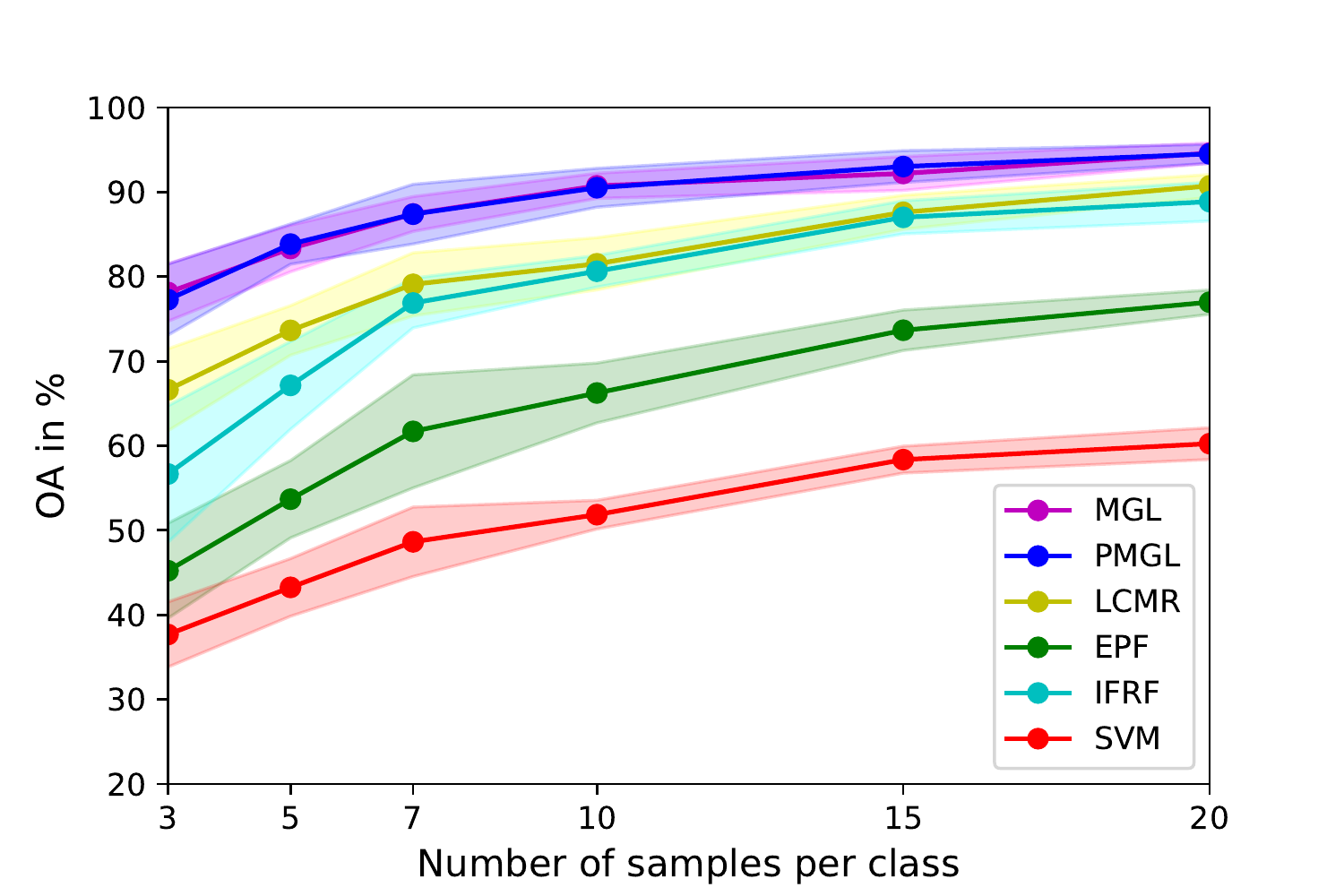}}
\subfloat[Pavia University]{  \includegraphics[width=2.2in]{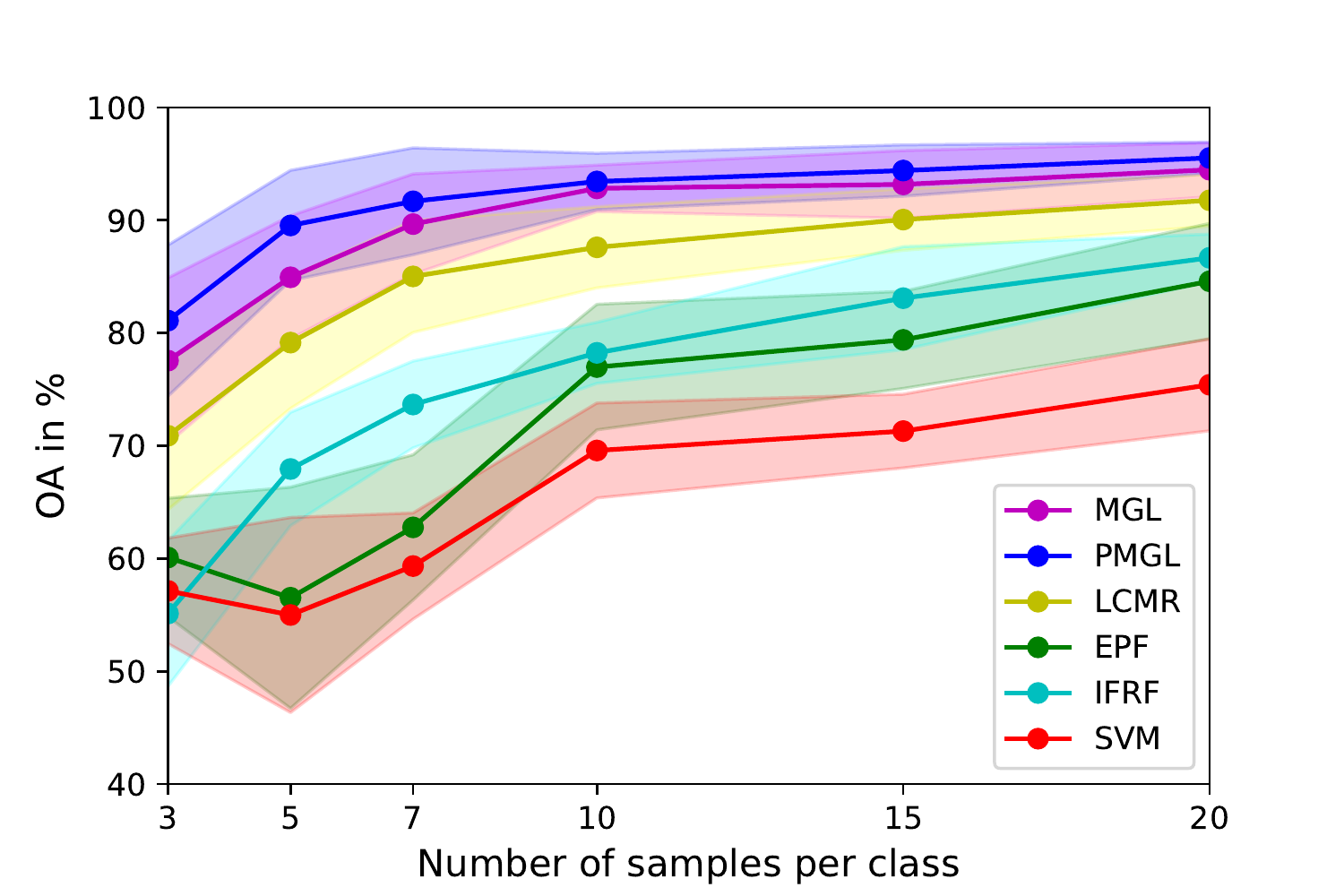}}
\subfloat[Salinas]{  \includegraphics[width=2.2in]{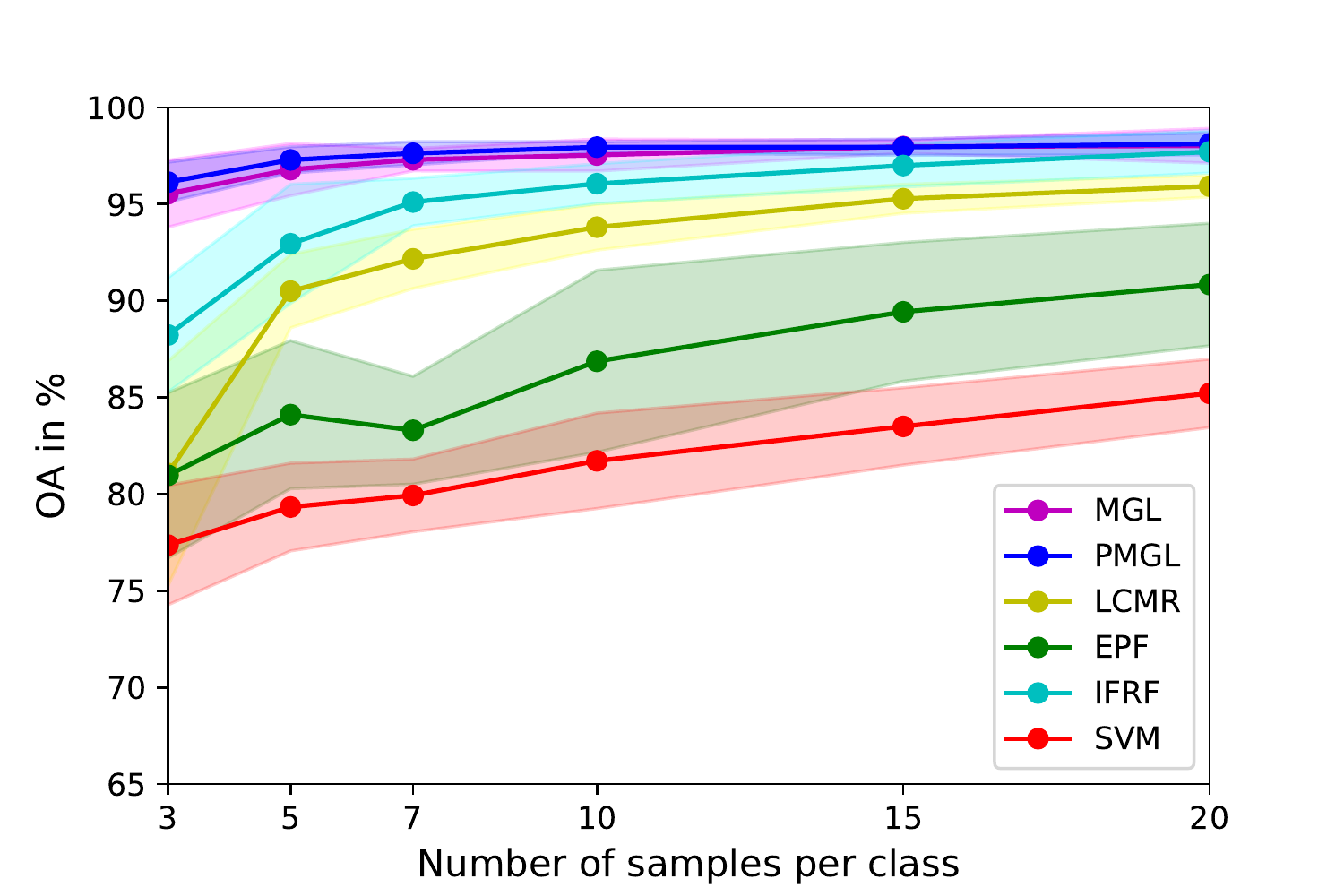}}

\subfloat[Indian Pines]{  \includegraphics[width=2.2in]{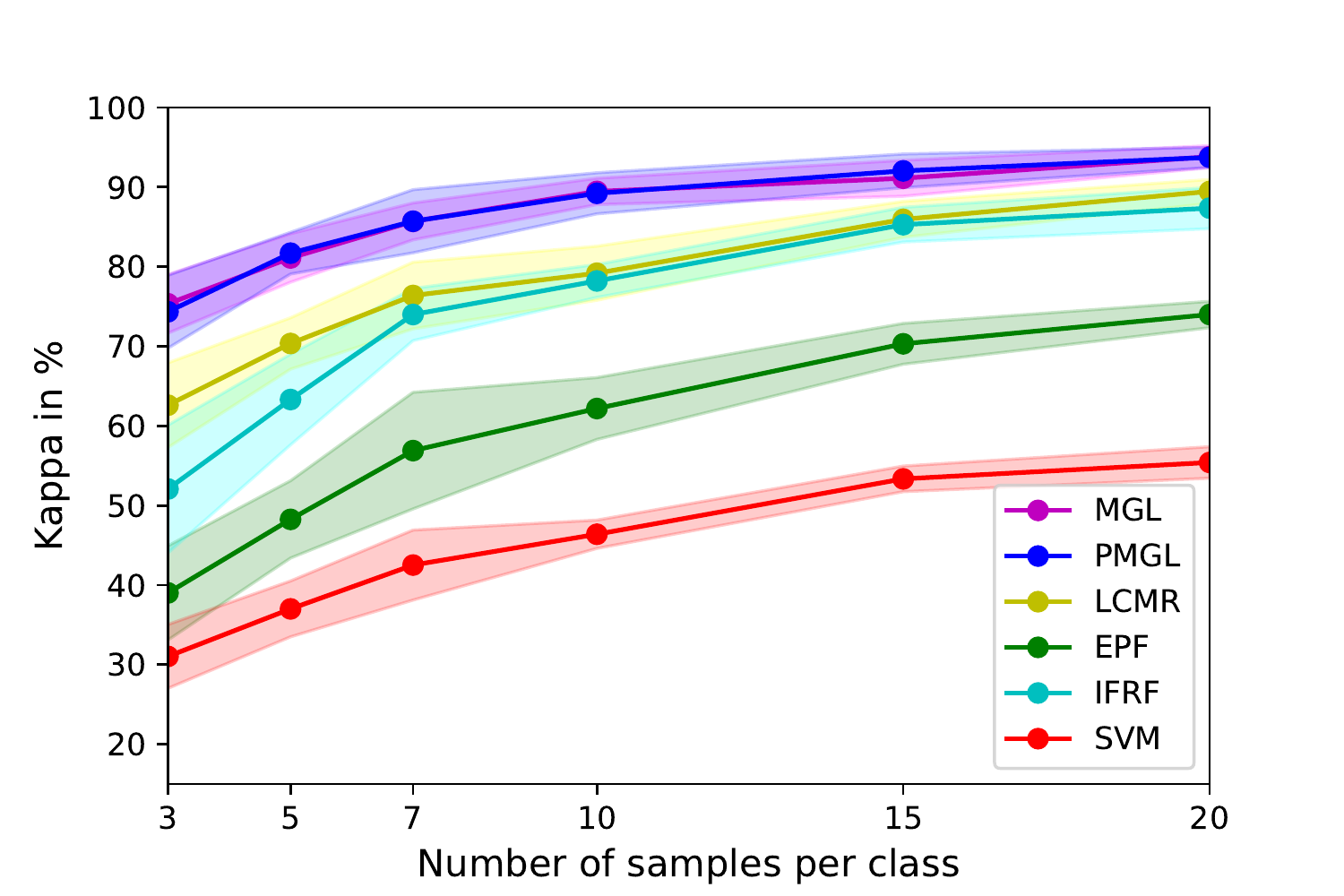}}
\subfloat[Pavia University]{  \includegraphics[width=2.2in]{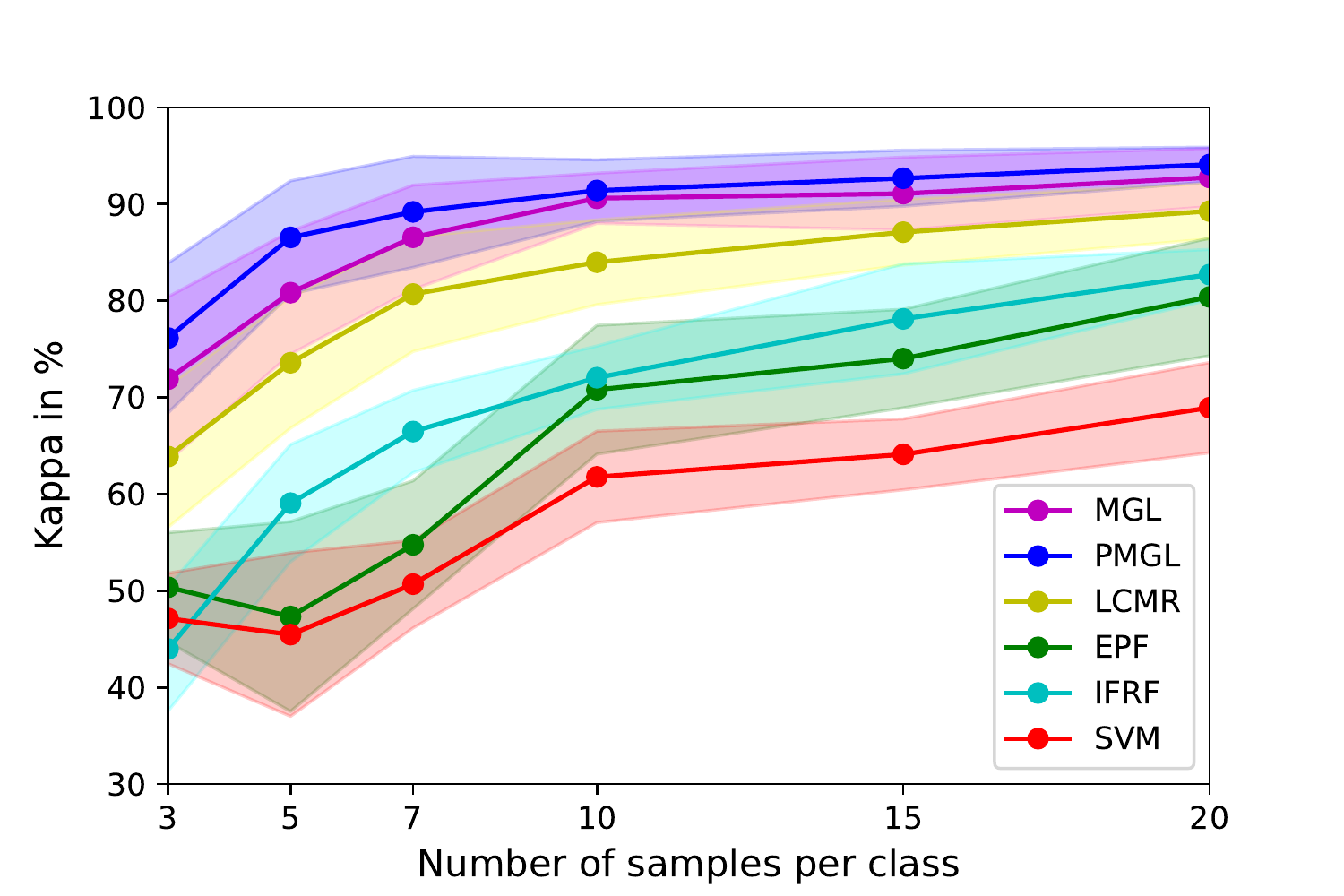}}
\subfloat[Salinas]{  \includegraphics[width=2.2in]{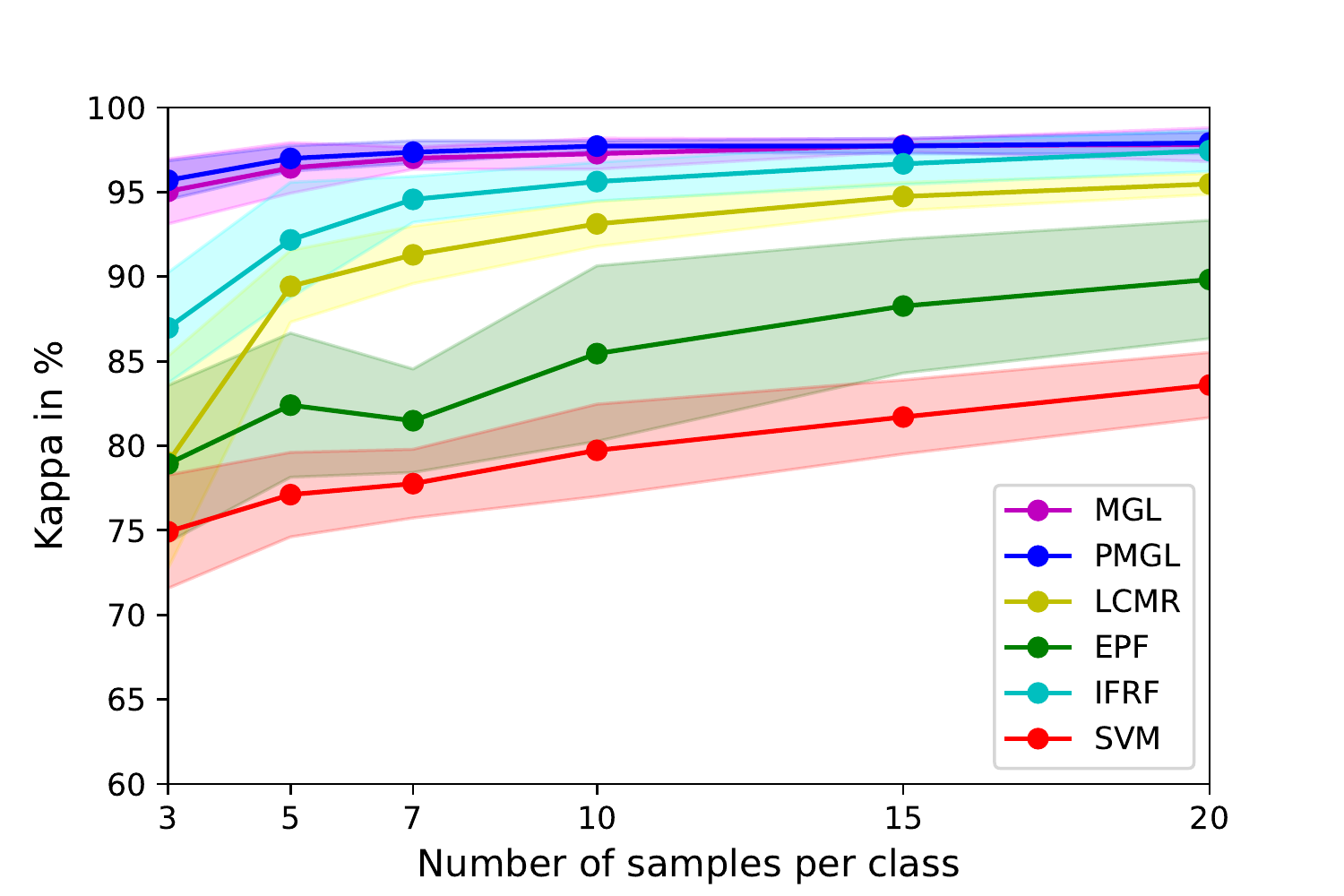}}
\caption{\footnotesize{Comparison of the classification accuracy (OA) and Kappa coefficient of different methods with varying number of training samples over 10 trials. The solid lines represent the mean while the shaded area covers the standard deviation from the mean.}}
  \label{figg1}
\end{figure*}

Performance is evaluated using a very limited amount of labels for training, with rates of $3-20$ samples per class, which are randomly selected, in several stages of experiments.
In the first instance, we conduct numerical evaluations and comparisons of classification accuracy of the proposed methods with state-of-the-art approaches, as detailed above, followed by an evaluation of the corresponding visual classification maps.
\\
{\bf E1}: We begin by evaluating the OA  and Kappa coefficient in comparison using a reduced label rate of 3-20 randomly selected labels per class, whose results are graphically displayed in Fig. \ref{figg1} 
for the three benchmark data sets. 
We observe that both proposed methods consistently outperform the other methods over the entire range of label rates for all three benchmark data sets. In particular, the performance gain is highest for lower label rates, signifying that the proposed pseudo-label guided graph-based methods perform strongly even when extremely few labels are available as a result of their superior model. LCMR and IFRF form the closest competitors for the Indian Pines and Salinas data sets with a maximum gap of approximately $10\%$ and $8 \%$ respectively to the closest competitor at 3 labelled samples, with LCMR being the dominant competitor for the more complex Pavia University data set with a gap of $10\%$ to PMGL.
Further, among the two proposed methods, in the lower label limit of the Pavia dataset, PMGL exhibits up to $4\%$ gain in OA performance, while for the other data sets, this gain is vanishingly small with the two methods performing comparably. This indicates that a more refined parameter selection can be beneficial for structurally complex data sets.

\begin{figure*}
\subfloat[]{ \includegraphics[width=2.4in]{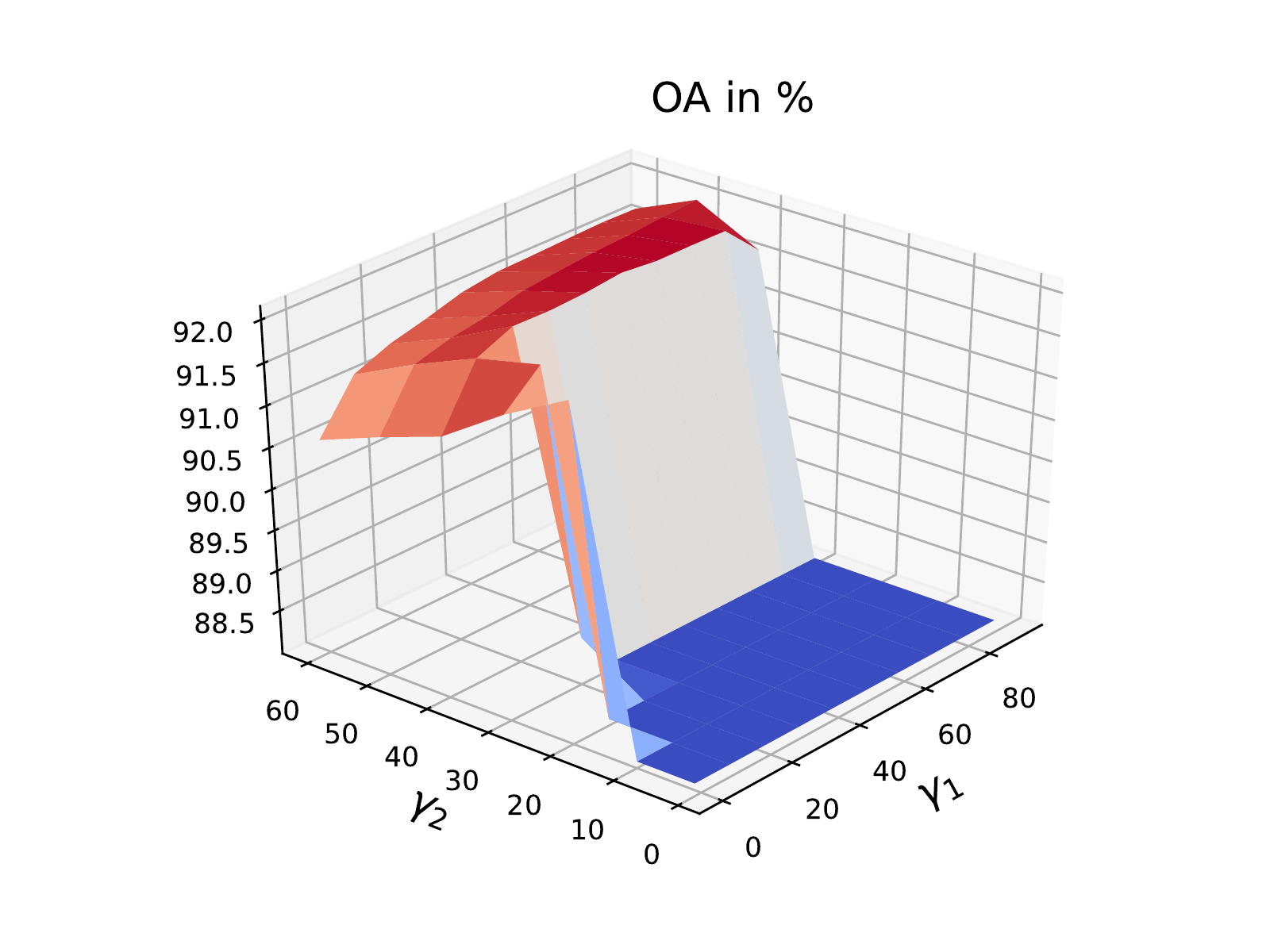}}
\subfloat[]{ \includegraphics[width=2.4in]{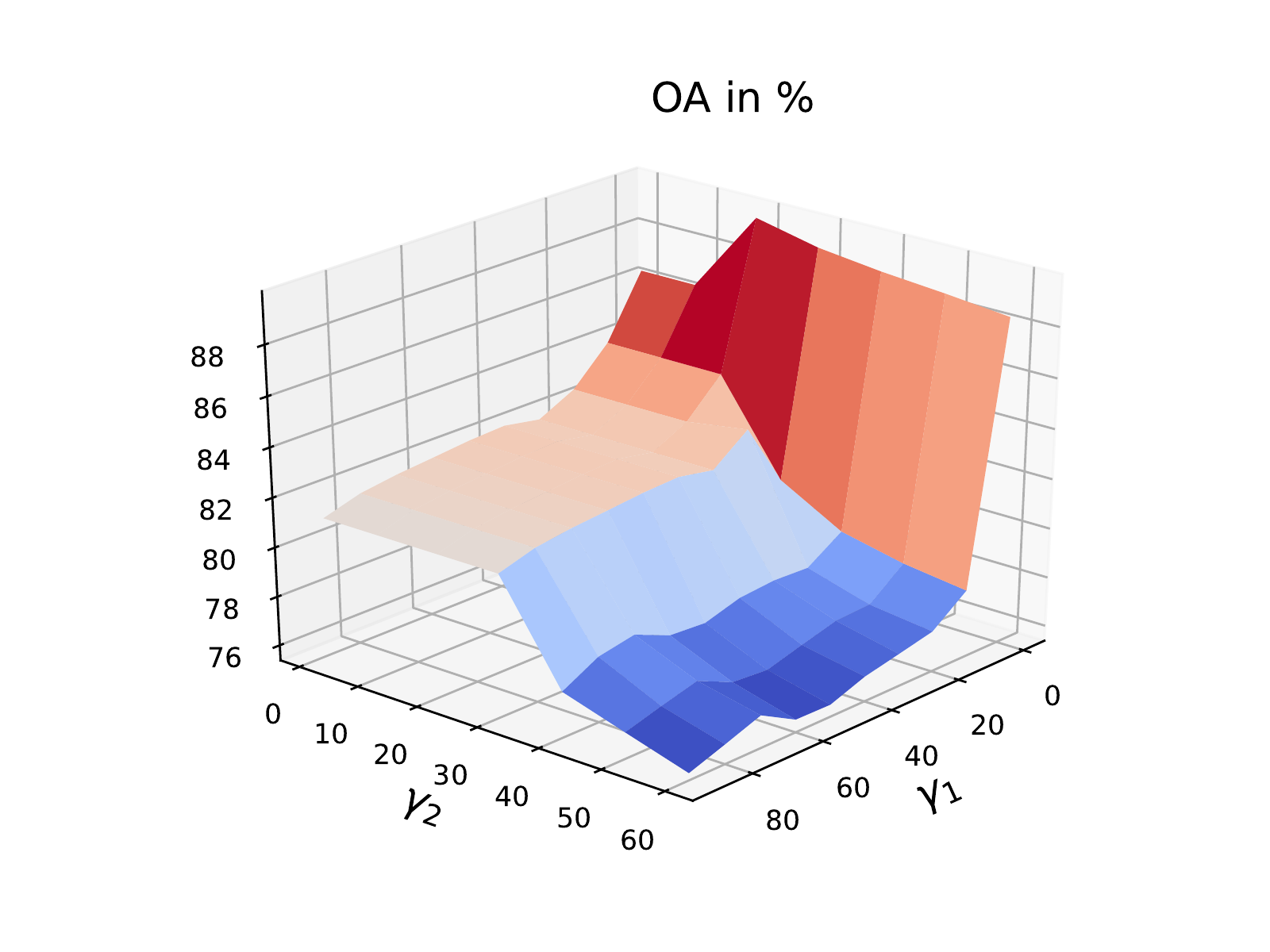}}
\subfloat[]{ \includegraphics[width=2.4in]{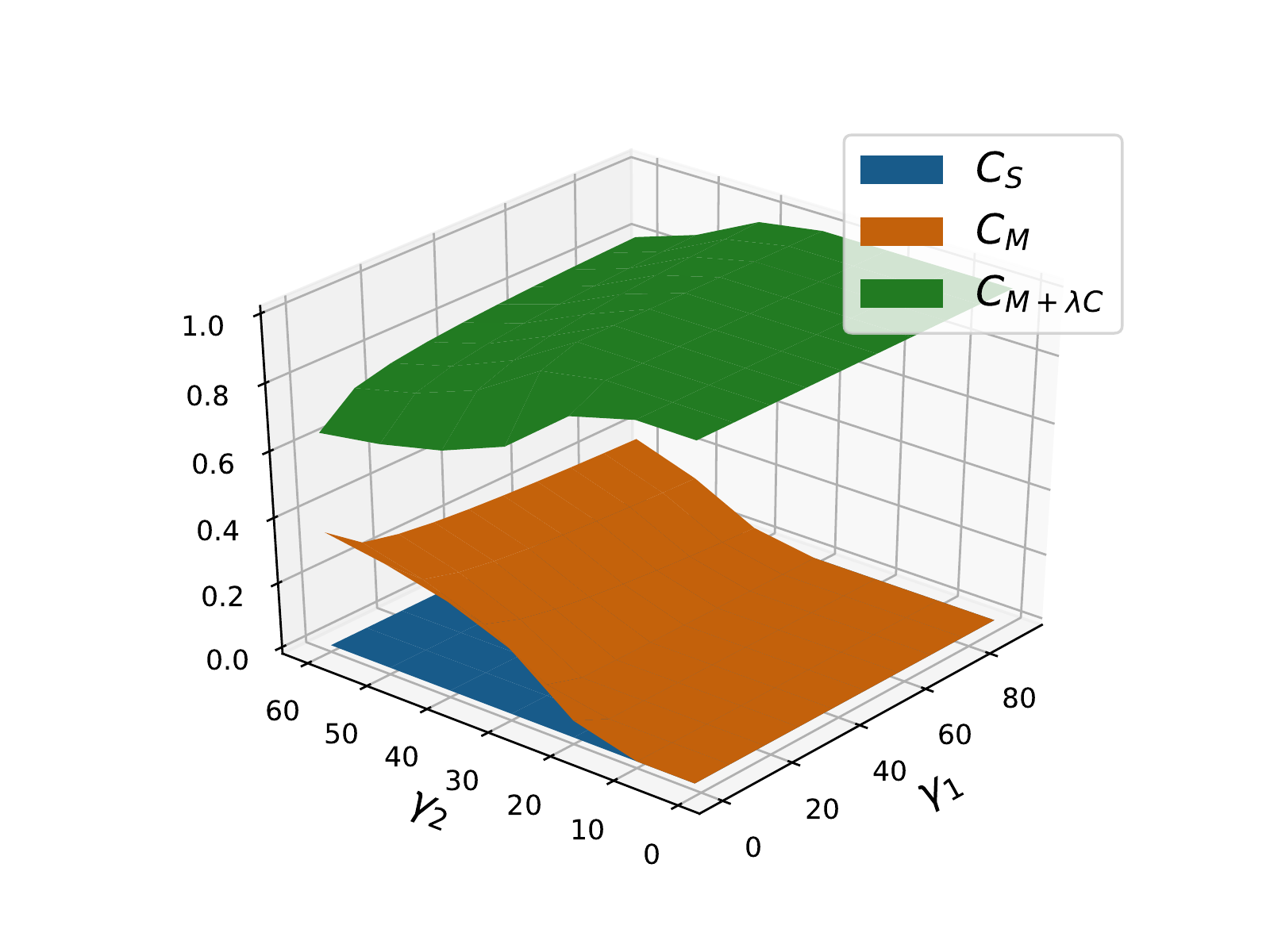}}
\caption{University of Pavia: results for 7 training samples per class, 3D plot of pseudo-label weight $\gamma_1$ vs feature regularization weight $\gamma_2$ against OA of PMGL with (a) $\gamma_3=0$, (b) $\gamma_3=1$, and (c) feature coefficient distribution for $\gamma_3=0$.}
  \label{figg2}
\end{figure*}

\begin{figure*}[!t]
\centering
\subfloat[Colour]{ \includegraphics[width=1.2in]{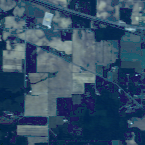}}\hfil
\subfloat[GT]{ \includegraphics[width=1.2in]{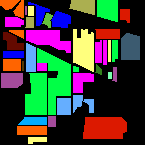}}\hfil
 \subfloat[MGL]{  \includegraphics[width=1.2in]{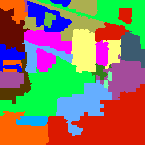}}\hfil
  \subfloat[PMGL]{\includegraphics[width=1.2in]{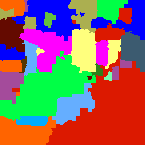}}\hfil
   \subfloat[LCMR]{ \includegraphics[width=1.2in]{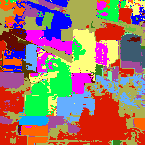}}\hfil
  \subfloat[EPF]{  \includegraphics[width=1.2in]{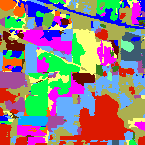}}\hfil
    \subfloat[IFRF]{  \includegraphics[width=1.2in]{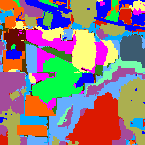}}\hfil
  \subfloat[SVM]{  \includegraphics[width=1.2in]{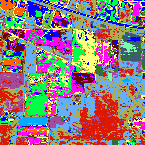}}\hfil
   \vspace{2mm}
 {\includegraphics[width=4.2in]{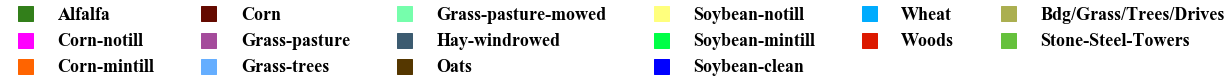}}
  \caption{Indian Pines: (a) Colour composite, (b) Ground truth, (c)-(h) classification maps produced using 7 labelled samples per class.}
  \label{ind_im}
\end{figure*}

\begin{figure*}[!t]
\centering
\subfloat[Colour]{  \includegraphics[width=0.84in]{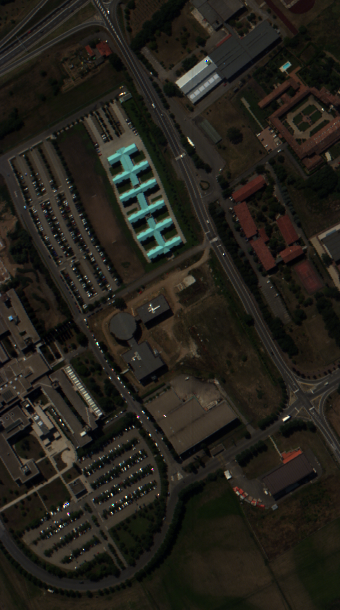}}\hfil
\subfloat[GT]{  \includegraphics[width=0.84in]{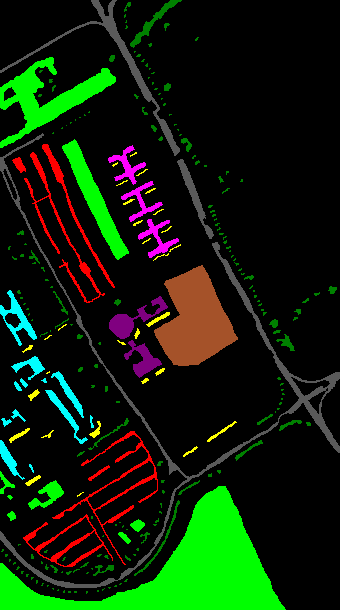}}\hfil
\subfloat[MGL]{   \includegraphics[width=0.84in]{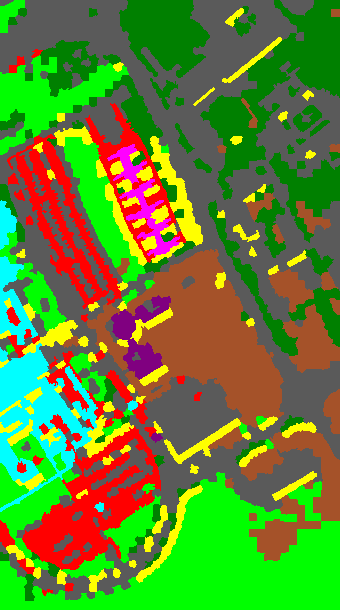}}\hfil
 \subfloat[PMGL]{ \includegraphics[width=0.84in]{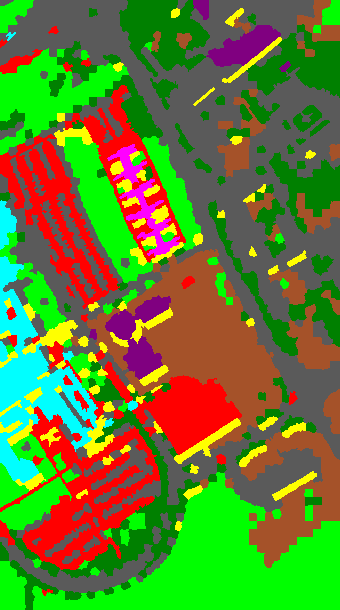}}\hfil
 \subfloat[LCMR]{  \includegraphics[width=0.84in]{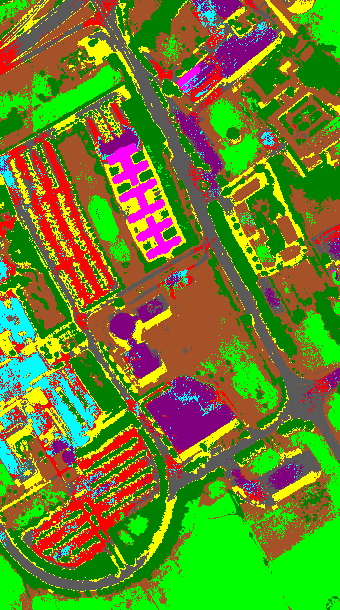}}\hfil
   \subfloat[EPF]{ \includegraphics[width=0.84in]{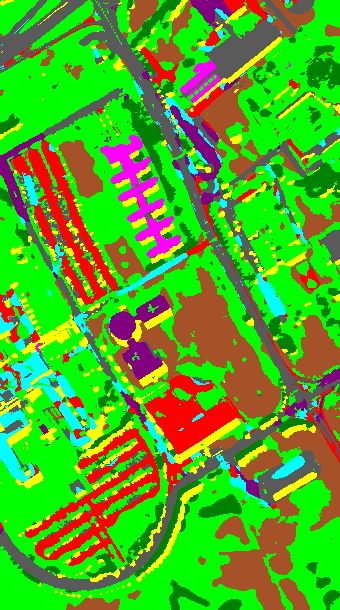}}\hfil
     \subfloat[IFRF]{ \includegraphics[width=0.84in]{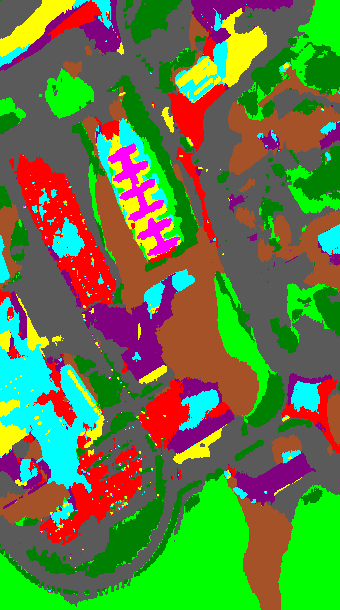}}\hfil
 \subfloat[SVM]{ \includegraphics[width=0.84in]{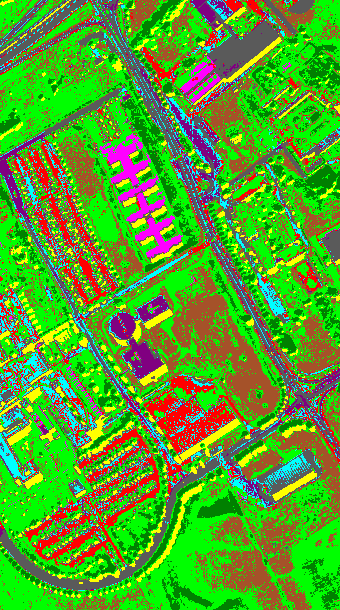}}\hfil
 \vspace{2mm}
 {\includegraphics[width=4in]{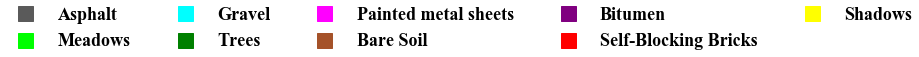}}
   \caption{Pavia University: (a) Colour composite, (b) Ground truth, (c)-(h) classification maps produced using 7 labelled samples per class.}
   \label{pu_im}
\end{figure*}

\begin{figure*}[!t]
\centering
\subfloat[Colour]{  \includegraphics[width=0.84in]{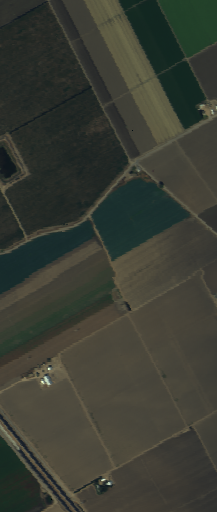}}\hfil
\subfloat[GT]{  \includegraphics[width=0.84in]{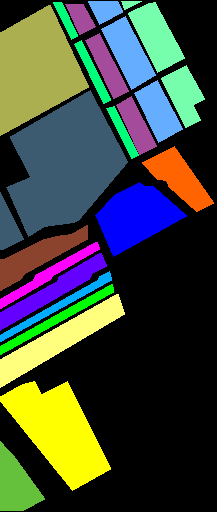}}\hfil
\subfloat[MGL]{  \includegraphics[width=0.84in]{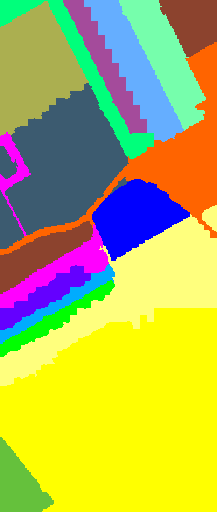}}\hfil
\subfloat[PMGL]{  \includegraphics[width=0.84in]{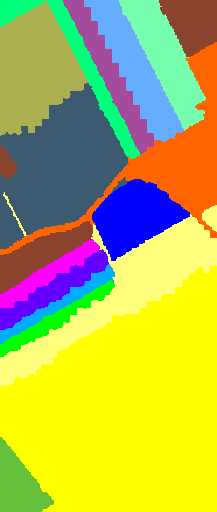}}\hfil
 \subfloat[LCMR]{ \includegraphics[width=0.84in]{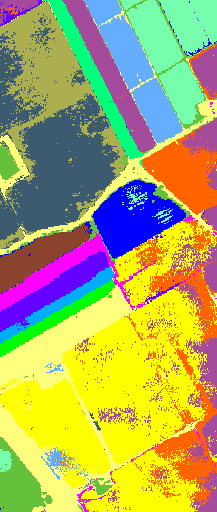}}\hfil%
 \subfloat[EPF]{ \includegraphics[width=0.84in]{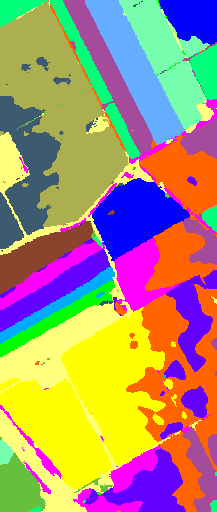}}\hfil
\subfloat[IFRF]{  \includegraphics[width=0.84in]{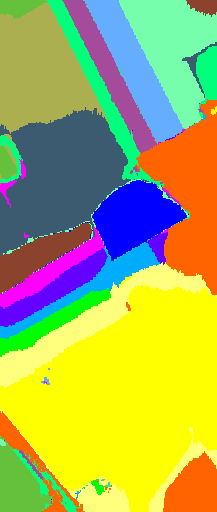}}\hfil
 \subfloat[SVM]{ \includegraphics[width=0.84in]{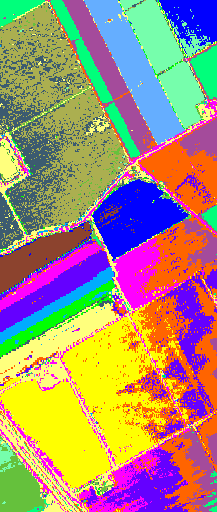}}\hfil
 \vspace{2mm}
 {\includegraphics[width=6in]{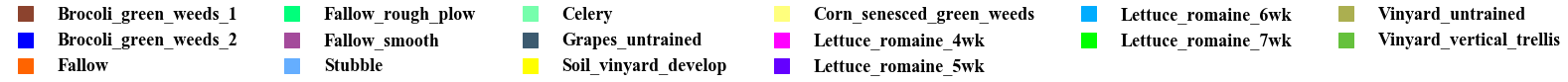}}
 \caption{Salinas: (a) Colour composite, (b) Ground truth, (c)-(h) classification maps produced using 7 labelled samples per class.}
  \label{sal_im}
\end{figure*}

\noindent {\bf E2}:
To demonstrate the influence of the pseudo-labels and feature coefficients in PMGL on performance, we consider the overall accuracy at a fixed label rate of 7 samples per class with varying feature parameters over 10 trials.
In particular, in order to illustrate the interaction between pseudo-label contribution and feature contribution, we fix pseudo-label parameter $\gamma_3$, while varying $\gamma_1$, along with the feature weight-regularization parameter $\gamma_2$, and consider the mean OA in a 3D plot. We report results for the University of Pavia data set, as the most structurally complex of the three, in Fig. \ref{figg2} with the OA plotted against $\gamma_1$ and $\gamma_2$ and $\gamma_3=0$ in (a), and $\gamma_3=1$ in (b), with (c) showing the resulting feature coefficient distribution for (a).
We observe that in (a), OA is highest when both $\gamma_1$ and $\gamma_2$ are increased, generating a perturbation toward more evenly distributed coefficients, which, as shown in (c), corresponds to the 
gradual matching of composite coefficient $c_{M+\lambda C}$ and spectral mean coefficient $c_{M}$, while the contribution of the spatial mean coefficient $c_{S}$ is negligible.
When additionally, the pseudo-label feature is incorporated into the graph via $\gamma_3$, the coefficient distribution for the best OA changes, instead overall moving toward one dominant superpixel feature.
While we observe interactions between pseudo-label and superpixel features which drive performance, the feature regularization parameter is ultimately dependent on the data set and the selected features at hand.
\\
\noindent {\bf E3}:
For each data set we use 7 labeled samples per class and run the methods again to calculate the OA, Kappa coefficient, AA, and a full class by class accuracy breakdown over 10 trials. For all three data sets, the proposed methods MGL and PMGL consistently outperform the competitors in OA, Kappa coefficient and AA, and for the majority of per class accuracies.
For the Indian Pines data set, PMGL performs only slightly better than MGL in the first three measures, and around $10 \%$ improvement in OA over its closest competitor, LCMR.
In the case of the University of Pavia data set, the gain of PMGL over MGL is even larger with 3-4$\%$, followed by LCMR with $10$$\%$ improvement in OA. Lastly, for the Salinas data set, MGL and PMGL perform comparativey well, with a gain of $3$$\%$ over their closest competitor, IFRF. Overall, we observe that while PMGL utilizes more intricate relations and selective feature contributions, MGL is still close in performance, with a gain of the former becoming more evident for increasingly complex data sets, such as Pavia University.
\begin{table*}[!t]
\caption{OA, Kappa, AA  and per class breakdown in $\%$ with 7 training labels per class}
\label{table1}
\centering
\begin{tabular}{|c|c|c||c|c|c|c|}
\cline{1-7}
\multicolumn{7}{|c@{}|}{{\bf Indian Pines}}\\
\hline
{CLASS} &          MGL &         PMGL &          LCMR \cite{lcmr} &            EPF \cite{epf}&          IFRF \cite{IFRF}&            SVM \cite{svm} \\
\hline
C1    &   $97.95\pm1.03$ &   $97.95\pm1.03$ &   ${\bf 99.23\pm1.24}$ &   $95.13\pm7.69$ &   ${\bf 99.23\pm1.24}$ &  $72.05\pm13.54$ \\
C2    &   $66.17\pm10.17$ &  $70.63\pm11.86$ &   ${\bf 73.36\pm8.24}$ &   $39.32\pm9.52$ &   $65.58\pm9.98$ &   $33.12\pm4.41$ \\
C3    &   $76.11\pm9.73$ &    ${\bf 86.12\pm8.26}$ &   $62.08\pm8.98$ &  $55.65\pm12.92$ &    $76.1\pm9.16$ &   $44.63\pm6.07$ \\
C4    &   ${\bf 96.17\pm4.01}$ &   $92.3\pm7.54$ &   $93.48\pm6.45$ &  $78.13\pm20.28$ &    $87.3\pm7.99$ &  $48.96\pm12.01$ \\
C5    &   ${\bf 88.84\pm9.09}$ &   $88.47\pm8.83$ &   $88.55\pm8.04$ &   $83.8\pm10.94$ &    $80.8\pm6.47$ &  $70.82\pm10.22$ \\
C6    &  ${\bf 94.54\pm10.67}$ &   ${\bf 93.43\pm8.9}$ &   $86.43\pm8.34$ &   $90.06\pm8.05$ &  $86.02\pm13.87$ &  $68.49\pm10.61$ \\
C7    &    ${\bf 100.0\pm0.0}$ &    ${\bf 100.0\pm0.0}$ &    ${\bf 100.0\pm0.0}$ &    $98.1\pm2.46$ &    ${\bf 100.0\pm0.0}$ &  $83.33\pm15.27$ \\
C8    &    ${\bf 100.0\pm0.0}$ &    ${\bf 100.0\pm0.0}$ &   $97.62\pm4.47$ &   $68.3\pm16.46$ &   $99.62\pm0.71$ &   $56.28\pm9.54$ \\
C9    &    ${\bf 100.0\pm0.0}$ &    ${\bf 100.0\pm0.0}$ &    ${\bf 100.0\pm0.0}$ &    ${\bf 100.0\pm0.0}$ &    ${\bf 100.0\pm0.0}$ &  $89.23\pm12.67$ \\
C10   &    ${\bf 88.05\pm7.41}$ &   ${\bf 89.97\pm7.04}$ &  $76.29\pm11.93$ &   $61.06\pm17.8$ &   $79.19\pm8.29$ &  $45.67\pm10.66$ \\
C11   &   ${\bf 87.53\pm6.94}$ &   ${\bf 82.07\pm10.0}$ &   $61.55\pm9.07$ &  $48.35\pm13.16$ &  $64.08\pm12.69$ &   $38.08\pm7.69$ \\
C12   &  ${\bf 86.64\pm12.78}$ &   ${\bf 93.16\pm10.15}$ &    $79.68\pm7.3$ &  $57.59\pm25.11$ &  $68.45\pm12.47$ &  $40.39\pm11.65$ \\
C13   &    ${\bf 99.49\pm0.0}$ &    ${\bf 99.49\pm0.0}$ &    $99.19\pm0.9$ &    ${\bf 99.49\pm0.0}$ &   $98.59\pm1.52$ &   $90.86\pm4.36$ \\
C14   &   ${\bf 97.02\pm8.58}$ &   $94.42\pm8.77$ &   $95.86\pm5.15$ &  $76.92\pm23.07$ &   $82.22\pm7.23$ &  $60.51\pm16.01$ \\
C15   &   ${\bf 94.51\pm9.39}$ &   ${\bf 94.64\pm9.29}$ &   $92.32\pm9.22$ &  $41.42\pm13.57$ &   $86.15\pm7.41$ &   $31.77\pm7.52$ \\
C16   &   $93.72\pm1.74$ &   $93.72\pm1.74$ &   ${\bf 99.19\pm2.57}$ &   $95.23\pm6.79$ &   $97.09\pm4.04$ &  $85.58\pm10.11$ \\
\hline
OA    &   ${\bf 86.75\pm2.31}$ &   ${\bf 86.93\pm2.79}$ &   $77.88\pm4.07$ &    $60.7\pm6.92$ &   $75.77\pm4.65$ &   $47.84\pm3.29$ \\
Kappa &   ${\bf  84.96\pm2.6}$ &   ${\bf 85.19\pm3.12}$ &   $75.14\pm4.51$ &   $56.16\pm7.59$ &   $72.84\pm5.06$ &    $42.07\pm3.6$ \\
AA    &   ${\bf 91.67\pm1.31}$ &   ${\bf 92.27\pm1.11}$ &    $87.8\pm2.27$ &   $74.28\pm4.98$ &   $85.65\pm2.77$ &   $59.99\pm3.48$ \\
\hline
\\
\cline{1-7}
\multicolumn{7}{|c@{}|}{{\bf University of Pavia}}\\
\hline
{CLASS} &          MGL &         PMGL &          LCMR \cite{lcmr} &            EPF \cite{epf}&          IFRF \cite{IFRF}&            SVM \cite{svm} \\
\hline
C1    &   $76.23\pm8.91$ &  $78.42\pm6.35$ &   ${\bf 79.92\pm9.14}$ &   $76.0\pm11.24$ &   $53.47\pm9.97$ &   $62.63\pm8.75$ \\
C2    &    ${\bf 92.53\pm7.9}$ &  ${\bf 96.12\pm5.36}$ &  $81.38\pm11.88$ &  $59.17\pm10.17$ &   $78.64\pm6.42$ &   $55.55\pm8.96$ \\
C3    &   ${\bf 87.5\pm10.43}$ &  ${\bf 90.19\pm8.26}$ &  $83.98\pm10.72$ &   $60.88\pm13.2$ &  $55.76\pm10.56$ &   $55.05\pm9.57$ \\
C4    &   $84.03\pm7.83$ &  $84.76\pm6.85$ &   ${\bf 93.72\pm4.43}$ &    $86.49\pm5.0$ &  $65.45\pm16.27$ &   $88.46\pm5.49$ \\
C5    &   $93.27\pm6.35$ & $93.68\pm5.89$ &    $97.2\pm7.54$ &  $96.76\pm10.23$ &   ${\bf 99.13\pm0.62}$ &   $96.63\pm8.49$ \\
C6    &   ${\bf 94.94\pm6.51}$ &  ${\bf 98.23\pm2.63}$ &   $84.82\pm8.41$ &  $63.46\pm13.17$ &    $83.4\pm7.23$ &   $57.34\pm9.54$ \\
C7    &   ${\bf 96.02\pm1.84}$ &  ${\bf 99.46\pm0.26}$ &  $85.15\pm12.58$ &   $94.32\pm9.97$ &  $77.41\pm14.24$ &   $86.22\pm8.86$ \\
C8    &  ${\bf 82.41\pm19.95}$ &  ${\bf 90.71\pm5.06}$ &    $71.03\pm9.2$ &  $74.57\pm15.13$ &   $71.8\pm10.57$ &  $64.32\pm10.18$ \\
C9    &   $92.96\pm3.37$ &  $95.9\pm2.03$ &   $94.46\pm2.06$ &   ${\bf 99.78\pm0.56}$ &   $50.54\pm10.7$ &   $99.66\pm0.58$ \\
\hline
OA    &   ${\bf 88.7\pm3.56}$ &  ${\bf 92.08\pm2.48}$ &   $82.58\pm5.74$ &   $68.81\pm4.31$ &   $72.63\pm3.09$ &   $63.15\pm3.64$ \\
Kappa &   ${\bf 85.3\pm4.44}$ &  ${\bf 89.61\pm3.11}$ &   $77.77\pm6.74$ &   $61.27\pm4.75$ &    $64.86\pm3.7$ &   $54.59\pm3.77$ \\
AA    &   ${\bf 88.87\pm2.35}$ &  ${\bf 91.94\pm1.6}$ &   $85.74\pm2.64$ &   $79.05\pm4.05$ &   $70.62\pm2.78$ &   $73.98\pm2.75$ \\
\hline
\\
\cline{1-7}
\multicolumn{7}{|c@{}|}{{\bf Salinas}}\\
\hline
{CLASS} &          MGL &         PMGL &          LCMR \cite{lcmr} &            EPF \cite{epf}&          IFRF \cite{IFRF}&            SVM \cite{svm} \\
\hline
C1    &   ${\bf 100.0\pm0.0}$ &   ${\bf 100.0\pm0.0}$ &  $99.89\pm0.21$ &    $99.77\pm0.5$ &  $98.23\pm5.61$ &   $97.85\pm1.26$ \\
C2    &   ${\bf 100.0\pm0.0}$ &  $ {\bf 100.0\pm0.0}$ &  $89.79\pm5.67$ &    $99.74\pm0.4$ &  $96.36\pm4.47$ &   $98.14\pm1.42$ \\
C3    &   ${\bf 100.0\pm0.0}$ &   ${\bf 100.0\pm0.0}$ &   $98.55\pm2.5$ &  $86.26\pm19.94$ &  $99.97\pm0.08$ &  $81.24\pm18.02$ \\
C4    &  $94.18\pm5.92$ &  $94.17\pm8.91$ &   ${\bf 100.0\pm0.0}$ &   $99.88\pm0.13$ &  $98.61\pm2.63$ &   $99.15\pm0.59$ \\
C5    &  $93.28\pm4.31$ &   $94.72\pm0.0$ &  $96.53\pm0.69$ &   ${\bf 97.62\pm1.81}$ &   $92.43\pm5.1$ &   $96.07\pm2.37$ \\
C6    &  ${\bf 99.59\pm0.07}$ &  ${\bf 99.59\pm0.06}$ &  $99.02\pm0.99$ &   $99.51\pm0.79$ &  $99.53\pm0.65$ &    $98.02\pm2.1$ \\
C7    &  ${\bf 99.89\pm0.06}$ &   ${\bf 100.0\pm0.0}$ &   $97.9\pm2.15$ &   $99.79\pm0.07$ &  $98.82\pm3.33$ &   $98.57\pm0.93$ \\
C8    &  ${\bf 98.45\pm0.71}$ &   ${\bf 97.46\pm1.84}$ &  $85.58\pm5.54$ &  $64.96\pm18.35$ &  $85.93\pm7.01$ &  $57.35\pm11.09$ \\
C9    &   ${\bf 100.0\pm0.0}$ &   ${\bf 100.0\pm0.0}$ &  $92.23\pm9.51$ &   $99.49\pm0.61$ &  $99.94\pm0.15$ &   $98.23\pm0.91$ \\
C10   &  $89.47\pm7.41$ &  $90.37\pm5.68$ &  ${\bf 97.44\pm1.37}$ &   $89.24\pm9.66$ &  $97.16\pm3.84$ &   $80.48\pm9.14$ \\
C11   &   $95.57\pm5.16$ &  $97.67\pm0.85$ &  ${\bf 99.91\pm0.08}$ &   $97.15\pm2.03$ &  $96.03\pm3.32$ &    $87.51\pm3.7$ \\
C12   &  $97.59\pm0.31$ &  $97.59\pm0.31$ &  $99.23\pm2.38$ &    ${\bf 100.0\pm0.0}$ &   $98.3\pm1.45$ &   $96.18\pm3.08$ \\
C13   &  $97.67\pm0.62$&  $97.48\pm0.36$ &  $98.26\pm0.85$ &   ${\bf 98.92\pm0.38}$ &  $95.73\pm5.84$ &    $98.22\pm0.6$ \\
C14   &  $92.91\pm4.8$ &  $94.65\pm2.95$ &  $93.13\pm5.84$ &    $95.3\pm1.66$ &  ${\bf 96.32\pm3.28}$ &   $88.86\pm2.69$ \\
C15   &  ${\bf 96.32\pm1.28}$ &  ${\bf 98.23\pm0.97}$ &  $82.71\pm7.54$ &  $71.74\pm21.97$ &  $93.98\pm3.76$ &  $59.05\pm12.78$ \\
C16   &   ${\bf 100.0\pm0.0}$ &   ${\bf 100.0\pm0.0}$ &  $91.23\pm7.49$ &   $94.71\pm4.17$ &  $98.12\pm3.02$ &   $92.33\pm4.68$ \\
\hline
OA    &   ${\bf 97.67\pm0.48}$ &   ${\bf 97.93\pm0.45}$ &  $91.99\pm1.34$ &   $87.14\pm3.22$ &   $94.89\pm1.7$ &   $81.98\pm1.54$ \\
Kappa &  ${\bf 97.41\pm0.54}$ &  ${\bf 97.7\pm0.5}$ &   $91.1\pm1.49$ &   $85.73\pm3.57$ &  $94.32\pm1.89$ &   $80.01\pm1.69$ \\
AA    &  ${\bf 97.18\pm0.69}$ &  ${\bf 97.62\pm0.65}$ &  $95.09\pm1.01$ &   $93.38\pm2.24$ &  $96.59\pm0.85$ &     $89.2\pm1.5$ \\
\hline
\end{tabular}
\end{table*}\\
\noindent {\bf E4}:
At last, we show the full classification maps produced for training with 7 samples per class for all methods in comparison. In Fig. \ref{ind_im}, classification maps for the Indian Pines data set illustrate increased smoothness and local homogeneity for the proposed graph-based MGL and PMGL, exemplifying their superiority. Their closest competitor, LCMR exhibits more noisy regions. 
Fig. \ref{pu_im} shows the University of Pavia classification maps, which as the most structurally complex scene of the three with some scattered classes, similarly exhibits smooth yet spatially refined classification results for the proposed MGL and PMGL, despite the inherent crudeness of superpixel segmentation and label regularization.
At last, for the Salinas data set in Fig. \ref{sal_im}, which due to its locally homogeneous regions and overall spatial regularity represents a simpler scene, the proposed methods still to manage to improve over existing methods, achieving near perfect classification.\\
It becomes evident that the use of superpixels, ensuring local homogeneity, as well as that of graphs, for refined local and global modelling, which incorporates pixel-and superpixel-level as well as spectral-spatial-label dependencies, facilitates a significant performance gain. Notably, EPF and IFRF employ a spectral-spatial filtering approach which can be likened to graph filtering, with the distinction that the latter is more flexible and versatile to model; nevertheless, the use of the SVM classifier in all competitors, as opposed to a graph-filter in the latter case, contributes to a comparatively decreased accuracy owing to the noise of the spectral-based SVM.

\section{Conclusion}
In this work, we have developed a pseudo-label-guided and superpixel-based graph learning framework for semi-supervised classification of HSI data. 
Specifically, we have presented two methods: while the former constructs a single dynamic graph by fusing different superpixel features along with a pseudo-label feature, the latter constructs a global graph as the sum of individual feature graphs whose contribution is informed through pseudo-label regularization. We have demonstrated on the basis of benchmark data sets the superiority in performance through quantitative and qualitative results in comparison to state-of-the-art methods, particularly in the small labelled sample limit.\\
The incorporation of multiple features as well as label information through pseudo-labels into the graph facilitate refined modelling of the complex dependencies present in HSI data, and ultimately leverage these for an improved classification performance.
Furthermore, the multi-stage workflow, which employs superpixels and a flexible, inherently sparse graph design with the option to reduce parameters through regularization, is versatile by allowing for multiple components and exploiting multiple levels of spectral-spatial and label-dependencies.
Nevertheless, it remains a challenge to completely eliminate parameters whilst maintaining competitive performance as well as to select the ideal features upon which the goodness of subsequent steps depends; ultimately, the process of feature selection/extraction and feature weighting presented is not exhaustive. In our future work, we wish to explore automated DL approaches for both feature selection and graph construction which can be guided by sophisticated model-based priors.


%



\section*{Acknowledgment}
The authors would like to thank Prof. D. Landgrebe from Purdue University and the NASA Jet Propulsion Laboratory for providing the hyperspectral data sets. They would also like to thank Prof D. Coomes from the Department of Plant Sciences, University of Cambridge, for his advice and support.
This research was carried out as part of the INTEGRAL project funded by GCRF (EPSRC) grant EP/T003553/1. In addition, CBS acknowledges support from the Philip Leverhulme Prize, the Royal Society Wolfson Fellowship, the EPSRC grants EP/S026045/1, EP/N014588/1, EP/T017961/1, the Wellcome Innovator Award RG98755, the Leverhulme Trust project Unveiling the invisible, the European Union Horizon 2020 research and innovation programme under the Marie Skodowska-Curie grant agreement No. 777826 NoMADS, the Cantab Capital Institute for the Mathematics of Information and the Alan Turing Institute.




\bibliographystyle{IEEEtran}
\bibliography{Bib2021IEEE}
%



%




\end{document}